\documentclass[]{bytedance_seed}

\DeclareUnicodeCharacter{1EA7}{\`a}
\DeclareUnicodeCharacter{1ED1}{\^o} 

\usepackage{xcolor} 
\usepackage{amssymb}

\newcommand{\dataset}{\textsc{DeR$^{2}$}} 
\newcommand{\settingZero}{Instruction-only }
\newcommand{\settingOne}{Concepts-only }
\newcommand{\settingTwo}{Related-only }
\newcommand{\settingThree}{Full-set }

\usepackage{natbib}

\usepackage{CJKutf8}

\usepackage{xargs}  

\usepackage{todonotes}  

\usepackage{multirow}

\usepackage{cleveref}

\usepackage{amsmath}
\usepackage{dsfont}

\usepackage{subcaption}

\usepackage{svg}
\usepackage{tcolorbox}

\usepackage[utf8]{inputenc}
\usepackage{booktabs}
\usepackage{graphicx}
\usepackage{array}
\usepackage{tabularx}
\usepackage{xcolor,colortbl}  
\definecolor{boxcolor}{HTML}{d92523} 
\definecolor{bulbcolor}{HTML}{e3b87f} 

\title{\centering Retrieval-Infused Reasoning Sandbox: A Benchmark for Decoupling Retrieval and Reasoning Capabilities}

\affiliation{ByteDance Seed, M-A-P}

\contribution{Full author list in Contributions}

\abstract{Despite strong performance on existing benchmarks, it remains unclear whether large language models can reason over genuinely novel scientific information. Most evaluations score end-to-end RAG pipelines, where reasoning is confounded with retrieval and toolchain choices, and the signal is further contaminated by parametric memorization and open-web volatility.
We introduce \dataset, a controlled deep-research sandbox that isolates document-grounded reasoning while preserving core difficulties of deep search: multi-step synthesis, denoising, and evidence-based conclusion making. \dataset decouples evidence access from reasoning via four regimes—Instruction-only, Concepts (gold concepts without documents), Related-only (only relevant documents), and Full-set (relevant documents plus topically related distractors)—yielding interpretable regime gaps that operationalize retrieval loss vs. reasoning loss and enable fine-grained error attribution. To prevent parametric leakage, we apply a two-phase validation that requires parametric failure without evidence while ensuring oracle-concept solvability. To ensure reproducibility, each instance provides a frozen document library (drawn from 2023–2025 theoretical papers) with expert-annotated concepts and validated rationales.
Experiments across a diverse set of state-of-the-art foundation models reveal substantial variation and significant headroom: some models exhibit mode-switch fragility, performing worse with the Full-set than with Instruction-only, while others show structural concept misuse, correctly naming concepts but failing to execute them as procedures.
}
\date{\today}

\correspondence{Ge Zhang at \email{zhangge.eli@bytedance.com}}
\checkdata[Project Page]{\url{https://retrieval-infused-reasoning-sandbox.github.io/}}

\begin{document}

\maketitle

\vspace{-3em}
\begin{figure}[h]
    \centering
    \includegraphics[width=.9\textwidth]{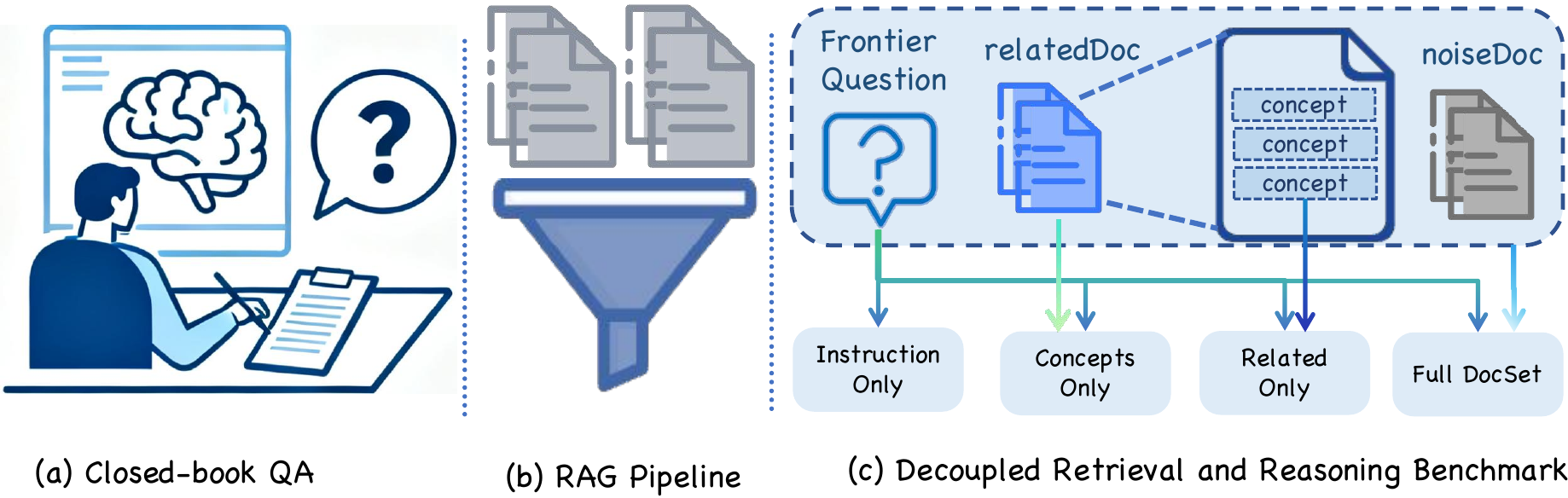}
    \caption{Closed-book QA measures intrinsic knowledge, while standard RAG entangles retrieval and reasoning—distractors can hide whether errors come from bad retrieval or failed evidence-based reasoning. \dataset decouples the two by evaluating the same question under controlled inputs (instruction/concepts/related/full), isolating failure causes.}
    \label{fig:teaser}
\end{figure}


\section{Introduction}
Recent foundation models have advanced from fluent assistants to strong problem solvers, achieving high performance on difficult, objectively graded tasks such as competition-style mathematics (e.g., AIME) and real-world software engineering (e.g., SWE-bench) \cite{patel2024aimeaioptimizationmultiple, jimenez2024swebenchlanguagemodelsresolve}. Meanwhile, agents—autonomous systems that navigate the internet to acquire and consolidate evidence—are increasingly used to extend models beyond parametric knowledge by coupling external search with internal reasoning and synthesizing information across multiple sources \cite{nakano2022webgptbrowserassistedquestionansweringhuman,ferrag2025llmreasoningautonomousai}. Within this broader shift, deep search has emerged as a foundational evaluation scenario: it stresses an agent’s ability to uncover hard-to-find facts or entities through multi-step exploration, denoising, and evidence-driven synthesis. Motivated by these demands, recent work has proposed stronger agentic designs \cite{li2025searcho1agenticsearchenhancedlarge,li2025webthinkerempoweringlargereasoning,sun2025zerosearchincentivizesearchcapability,jiang2025s3dontneeddata,li2025websailornavigatingsuperhumanreasoning} and introduced more challenging benchmarks that better capture deep-research behavior, including BrowseComp for hard-to-find web browsing tasks and HLE for frontier, expert-level closed-ended questions \cite{wei2025browsecomp,phan2025hle}.

However, we find that current deep-research evaluation practice suffers from three major limitations. \textbf{(1) End-to-end pipeline confounding with weak diagnosis:} many benchmarks score the full stack—retrieval, reranking, summarization/compression, context stitching, and reasoning—so variance is often dominated by toolchain choices rather than the model’s latent research reasoning capability. Moreover, most setups expose only a binary “correct/incorrect” outcome, which makes it difficult to attribute errors to distinct failure types such as missing the key concept, identifying the concept but failing to apply it, or being diverted by topically related noise that shifts the reasoning starting point and triggers mode-switch derailment. As a result, these evaluations provide limited guidance for model selection, debugging, or targeted training. \textbf{(2) Parametric leakage / memorization:} if tasks can be solved via memorized facts, broad encyclopedic knowledge, or easily guessable numeric forms, then correctness does not reliably indicate that the model consulted the crucial evidence and reasoned with it, contaminating the evaluation signal. \textbf{(3) Web volatility and irreproducibility:} benchmarks that depend on open-web retrieval inherit non-stationarity across time, region, indexing updates, and page instability (movement, deletion, or content changes), producing high variance across repeated runs and hindering controlled comparisons and ablations \cite{wei2025browsecomp}.

To address these limitations, we propose \textsc{\dataset}, a controlled research sandbox that isolates document-grounded reasoning while preserving the core difficulties of deep search: multi-step synthesis, noise filtering, and evidence-based conclusion making. We mitigate \textbf{(1) pipeline confounding and weak diagnosis} by decoupling evidence access from reasoning via four evaluation regimes: \textbf{Instruction-only} (parametric knowledge), \textbf{Concepts} (instruction + gold concept set; concept composition/scheduling without document extraction), \textbf{Related-only} (instruction + only relevant documents; extraction and reasoning under clear evidence), and \textbf{Full-set} (relevant documents plus topically related distractors; denoising and reasoning). This design yields interpretable regime-to-regime gaps that operationalize “retrieval loss” versus “reasoning loss,” and it supports error attribution, including missing concepts, concept misuse, and noise-induced mode switching. We mitigate \textbf{(2) parametric leakage} through a rigorous two-phase validation protocol: each problem must exhibit parametric failure (low accuracy without documents) while remaining document-solvable (a correct solution is achievable when provided oracle concepts). This jointly enforces novelty (not answerable from memorized knowledge alone) and tractability (answerable given the intended evidence). Finally, we address \textbf{(3) web volatility} by replacing open-web dependence with a frozen, per-instance document library (mean: 6.5 documents) drawn from 2023--2025 theoretical papers. Each library includes both necessary evidence and carefully selected distractors that mimic realistic, imperfect literature gathering. Each instance further provides expert-annotated concept sets and validated chain-of-thought rationales, enabling process-level evaluation of evidence usage rather than relying solely on final-answer accuracy.

Our experiments reveal two consistent failure patterns. \textbf{(1) Mode-switch fragility:} Some models perform better under Instruction-only than under Full-set, indicating that additional documents can disrupt the controller responsible for shifting from parametric reasoning to evidence-grounded reasoning, thereby degrading performance. \textbf{(2) Structural concept misuse:} even when models identify and restate the correct concepts, they often fail to execute them as procedures (e.g., applying rules step-by-step or instantiating abstractions), instead reverting to generic heuristics that appear plausible but produce incorrect conclusions.

Our contributions are threefold:
\begin{itemize}
\item We articulate desiderata for evaluating research-oriented LLMs and show that prior QA, RAG, and reasoning benchmarks do not sufficiently isolate evidence-grounded synthesis.
\item We release \dataset, a rigorously curated benchmark featuring frozen document libraries with embedded noise, expert concept annotations, and validated CoT rationales drawn from frontier scientific literature.
\item We provide a systematic analysis of leading LLMs under this controlled setup, revealing substantial headroom even for top-tier systems and identifying concrete failure modes that future architectures and training regimes must address.
\end{itemize}

\section{Benchmark}
\label{sec:benchmark}

\subsection{Benchmark Overview}
\label{subsec:benchmark_overview}
We introduce \dataset, a controlled benchmark for evaluating \emph{retrieval-infused reasoning} in scientific problem solving. The benchmark is designed to expose failure modes that are difficult to diagnose in end-to-end retrieval-augmented generation (RAG) or agent-based deep research evaluations.

\begin{figure*}[t]
    \centering
    \includegraphics[width=\textwidth]{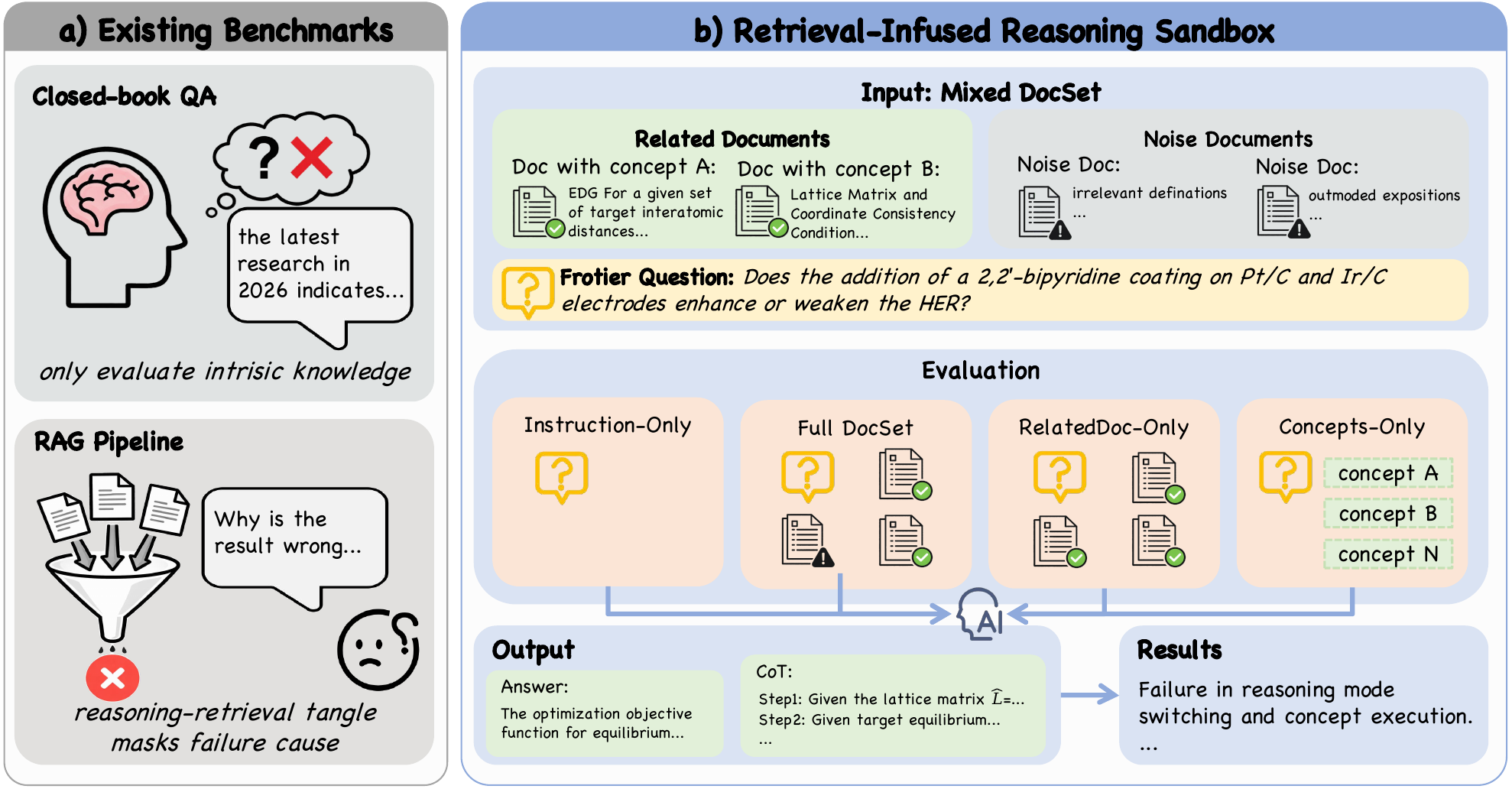}
    \caption{\textbf{(a) Existing benchmarks.} Closed-book QA only evaluates intrinsic (parametric) knowledge. RAG pipelines couple retrieval and reasoning end-to-end, creating a reasoning–retrieval tangle that masks the true cause of failure. \textbf{(b) Decoupled Retrieval and Reasoning Benchmark.} \dataset provides a controlled setting that decouples retrieval from reasoning by comparing Instruction-only, Full DocSet (related + noise), Related-only, and Concepts-only conditions, enabling diagnosis of grounding, mode-switching, and concept execution errors.}
    \label{fig:overview}
\end{figure*}

\dataset explicitly separates (a) a model's ability to \emph{identify and extract} solution-relevant concepts from a document set, from (b) its ability to \emph{compose and schedule} these concepts to derive the correct answer. This decomposition enables controlled diagnosis of where performance collapses in retrieval-infused reasoning.

Compared to conventional RAG benchmarks, \dataset provides (i) a controlled document set with both relevant and distractor documents, (ii) concept-level supervision that supports capability isolation, and (iii) a verifiability protocol ensuring that problems are not solvable by parametric memorization alone, but become solvable once the right concepts are provided.

\subsection{Task Definition}
\label{subsec:task_definition}

Each benchmark example asks the model to answer a frontier scientific question by grounding its reasoning in a provided set of documents. The target capability is not merely recalling definitions, but performing multi-step derivations that require selecting, instantiating, and coordinating multiple concepts.

Each example is defined by a tuple:
\begin{itemize}
    \item \textbf{Instruction}: A high-level academic question derived from cutting-edge scientific papers. The question should be solvable only if the model is provided with the correct concepts, making it impossible to answer without the relevant knowledge.
    \item \textbf{Concepts}: All the concepts, theorems, or formulas necessary to solve the instruction. These should not be overly specific experimental data, but rather generalizable scientific knowledge.
    \item \textbf{CoT (Chain of Thought)}: The reasoning process that connects the concepts to the final answer, illustrating how the model should use the provided knowledge step by step.
    \item \textbf{Answer}: The answer to the instruction, which could be a product, scalar value, or conclusion derived from the concepts and reasoning.
    \item \textbf{Doc Set}: A collection of documents that provide the necessary concepts and theories to answer the question. These documents include noise documents that are unrelated to the task, ensuring a balanced challenge for the model.
\end{itemize}

\dataset emphasizes \emph{research-style} reasoning, where the model must transform theoretical knowledge from documents into actionable steps for solving complex scientific problems. The task focuses on \emph{concept operationalization} (e.g., theorem instantiation, algorithm execution) rather than simply summarizing surface-level content.

\begin{figure*}[t]
    \centering
    \includegraphics[width=\textwidth]{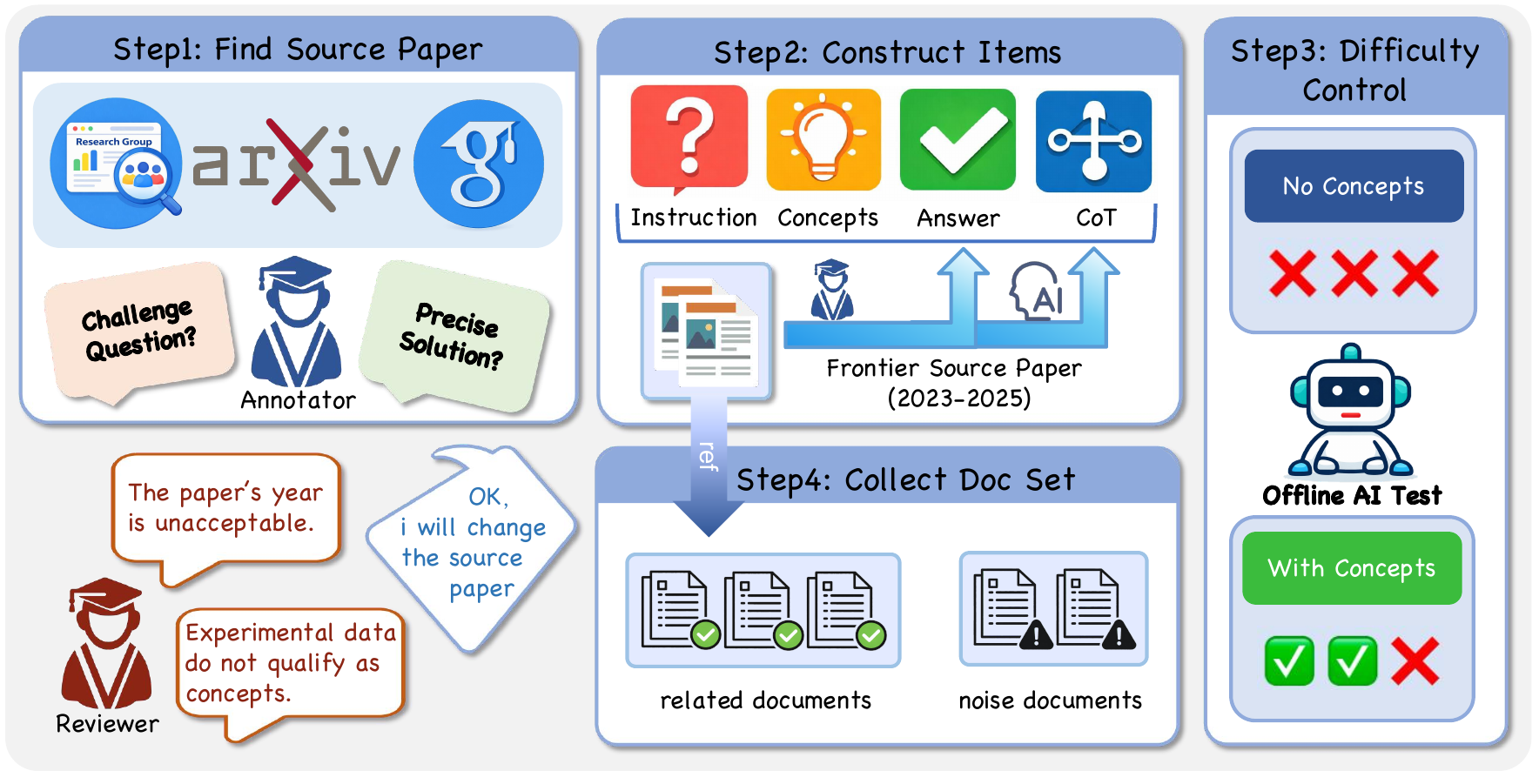}
    \caption{\textbf{Pipeline for constructing \dataset data:} PhD annotators gather theoretical papers from academic platforms, extracting instructions, concepts, CoT, and concise answers. These questions are designed to be challenging and represent cutting-edge academic problems. During the difficulty control phase, offline AI models must fail to answer correctly three times without concepts, but have a chance to answer correctly when provided with the relevant concepts. When the difficulty control conditions are satisfied, retrieve the related documents and noise documents from the references of the source paper. Finally, the reviewer conducts logical quality inspection and format quality inspection on the submitted data, and verifies the difficulty control conditions again.}
    \label{fig:building}
\end{figure*}

\subsection{Data Collection}
\label{subsec:dataset_and_scope}

We recruited annotators and reviewers from a highly selective pool: all contributors are current PhD students enrolled at top-tier Chinese universities (Project 985). To reduce domain-mismatch errors, both annotators and reviewers were strictly restricted to working \emph{only} within their own specialized academic fields. To attract top-tier annotators, a competitive compensation package was offered, with each task rewarded at a rate of 2500 RMB per annotated question. Reviewers were compensated at 350 RMB per audited item. A total of 81 annotators across various scientific disciplines participated in the data collection process.

\textbf{Step 1: Source paper acquisition and screening}.
Each annotation begins with collecting a \emph{source paper} published in \textbf{2023 between 2025}. We place special emphasis on provenance: when a paper has an arXiv preprint, we use the arXiv URL date as the publication timestamp for eligibility. We exclude \emph{applied} papers---including but not limited to physics/chemistry experiments, social surveys, and numerical simulations---because their conclusions are often primarily experimental and do not yield instructions solvable via concept-driven logical derivation. Eligible source papers must be \emph{theory-oriented}, spanning (i) theoretical foundational disciplines, (ii) theoretical engineering disciplines, and (iii) theory-driven interdisciplinary fields. As we know, the general workflow of annotators involves selecting source papers by tracking leading research groups in each domain and referencing materials published by prominent professors in the field.

\begin{figure}[h]
    \centering
    \includegraphics[width=.8\textwidth]{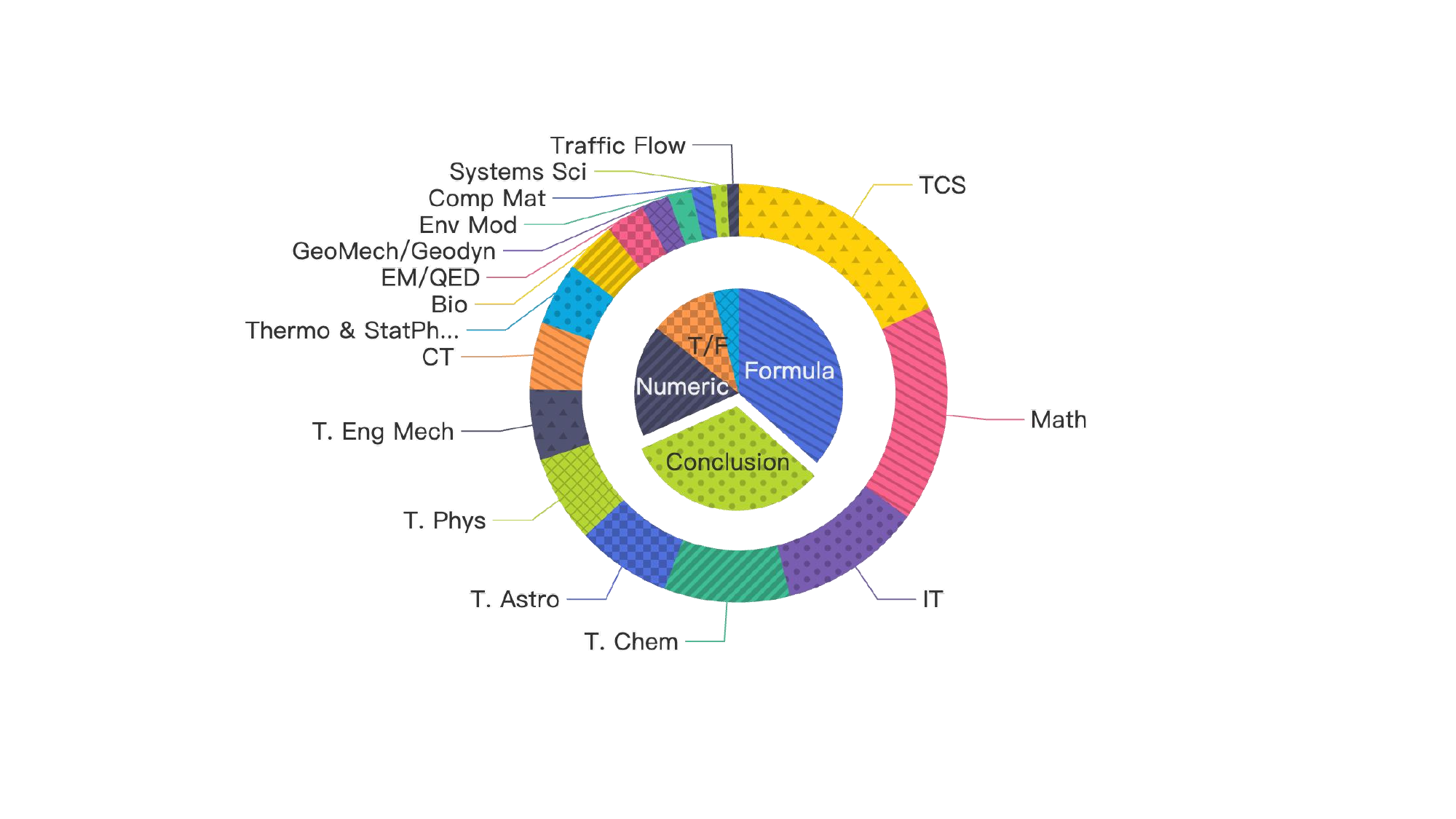}
    \caption{Distribution of problem domains and answer types, showing the disciplinary coverage of the benchmark and the structural diversity of target answers.}
    \label{fig:pie}
\end{figure}

\textbf{Step 2: Constructing the quadruple}(Instruction, Answer, Concepts, CoT).
After an annotator identifies an eligible source paper via abstract-level screening, they read the full paper to understand the problem setting, assumptions, and core argument. They then construct:
(i) an \textbf{Instruction} aligned with the paper's primary research question, with a scope that is neither overly broad (admitting multiple valid answers) nor overly narrow (only true within the specific paper instance);
(ii) a complete \textbf{Answer} that directly resolves the Instruction, prioritizing a short \emph{artifact}, \emph{scalar}, or \emph{concise conclusion}. If the Answer is necessarily long, the annotator must provide a rigorous \textbf{checklist} of key scoring items to make verification objective;
(iii) a set of \textbf{Concepts} capturing all and only the necessary theoretical concepts/theorems/formalisms required to solve the Instruction (not incidental experimental measurements);
(iv) a step-by-step \textbf{CoT} (reasoning trace) from Instruction to Answer, derived from the paper's theoretical argumentation, ensuring that every knowledge item used in the CoT is included in the Concepts list.

Annotators may use AI tools for \emph{brainstorming} or summarization, but the Instruction and Answer must be finalized by the annotator after fully understanding the paper. We explicitly observed two recurring AI failure modes: (a) selecting peripheral intermediate quantities as the Answer, yielding an Instruction that only holds inside the specific paper context; (b) generating overly open-ended Instructions where the provided Answer becomes only one of many plausible solutions. Therefore, human authorship is required for the final Instruction--Answer pair.

\textbf{Step 3: Difficulty calibration under offline LLM testing}.
To ensure that each item is sufficiently frontier and non-trivial, we perform controlled difficulty checks using \emph{offline} models (e.g., DeepSeek-R1-0528 or the Doubao web client with networking disabled). We run each condition three times and record correctness:
\begin{itemize}
    \item \textbf{\settingZero} the model answers the Instruction without Concepts. All three attempts must be incorrect.
    \item \textbf{\settingOne} the model answers with Concepts provided. Across three attempts, it must be correct at least once and incorrect at least once.
\end{itemize}
If the model is correct in all three Instruction+Concepts trials, we apply one of two remedies. Option A adds a \emph{deep validation} step by replacing Concepts with a document set (Related + Noise) and retesting; if the model remains always-correct, the item is discarded, whereas if it fails at least twice, the item can be retained (this option is discouraged when the doc set exceeds three PDFs due to context limitations in common LLM interfaces). At this point, option B can still be adopted, whereby annotators can iteratively edit the Instruction and/or Concepts (by adding/removing concepts or narrowing the question scope) to increase the difficulty until the calibration criteria are satisfied.

\textbf{Step 4: Document set construction }.
Once the quadruple passes difficulty calibration, the annotator constructs the document set. Starting from the source paper's references, the annotator locates at least one \textbf{Related} document for each Concept (a single related document may cover multiple Concepts). The annotator then curates several \textbf{Noise} documents that are topically adjacent to the Instruction but do not contribute solution-relevant concepts. For all documents, the annotator must verify: (i) Related documents contain the needed Concepts; (ii) Noise documents do not contain any solution Concepts while remaining on-topic; and (iii) \emph{no} document contains the Answer verbatim or in an equivalent directly-recoverable form.

\textbf{Review and quality assurance }(QA).
Reviewers are selected from annotators with \textbf{at least three accepted annotations}. Each reviewed item is checked for both format compliance and scientific validity: (i) the Instruction is a frontier academic question grounded in a 2023--2025 theory-oriented source paper, self-contained, and appropriately scoped; (ii) the Answer is unique, unambiguous, and easy to verify (or accompanied by a strict checklist when long); (iii) the Concepts are \emph{necessary} knowledge items for solving the Instruction, contain no irrelevant content, and are theoretical constructs rather than paper-specific experimental values; (iv) the CoT is a coherent concept-grounded derivation that uses \emph{all} Concepts (no omissions), does not rely on experimental measurements as premises, and yields a unique conclusion consistent with the Answer; and (v) the difficulty calibration tests are re-run by the reviewer and the document set is re-audited to confirm that no document leaks the Answer.

\paragraph{Collaboration loop.}
Annotators and reviewers collaborated in a shared online document workspace with high-frequency feedback. This iterative workflow enabled the rapid correction of ambiguity, scope issues, concept leakage, and document contamination, thereby improving consistency and ensuring that all accepted items met the benchmark’s structural and scientific requirements.

\section{Experiments}

\subsection{Evaluation Settings}
\label{subsec:eval_settings}

We evaluate models under four controlled input configurations that explicitly decouple \emph{retrieval capability} (selecting and extracting solution-relevant evidence from long contexts) from \emph{reasoning capability} (composing and scheduling the extracted concepts to derive the final answer). The key idea is to provide the model with progressively more information — ranging from none to oracle concepts, to clean evidence, and finally to evidence mixed with distractors — so that performance gaps can be attributed to specific failure sources rather than being conflated in an end-to-end score. 

We consider the following four settings, each evaluated independently (see \ref{tab:settings}).

These settings enable fine-grained diagnosis of three recurring failure sources:
(i) \textbf{Knowledge loss} (\settingOne $\rightarrow$ \settingZero gap),
(ii) \textbf{Retrieval loss} (\settingTwo $\rightarrow$ \settingOne gap), 
and (iii) \textbf{Noise-induced loss} (\settingThree $\rightarrow$ \settingTwo gap). To ensure the decoupling is meaningful, we keep the input format and evaluation procedure identical across settings; only the provided information differs.

\subsection{Models and Evaluation Protocol}
\label{subsec:models}

We evaluate a diverse set of state-of-the-art models. Concretely, our evaluated pool includes 
GPT-5.1~\cite{openai2025gpt51}, 
Gemini-3-Pro~\cite{google2025gemini3pro} and Gemini-2.5-Pro~\cite{comanici2025gemini}, 
Claude-Opus-4.1~\cite{claude-opus-4-1} and Claude-Sonnet-4.5~\cite{claude4.5}, 
DeepSeek-V3.1 and DeepSeek-V3.2~\cite{deepseek2024v32}, 
Doubao-1.6 and Doubao-1.8~\cite{bytedance2024doubao}, 
Qwen-3-max~\cite{yang2025qwen3}, 
Moonshot-Kimi-k2-thinking~\cite{kimiteam2025kimik2openagentic}, 
and GLM-4.6~\cite{zhipu2025glm46}. 
We aim to evaluate the retrieval-enhanced reasoning capabilities of different models in a controlled sandbox environment, rather than optimizing for any specific vendor stack.

\paragraph{Controlled evaluation interface.}
All models are evaluated under the same protocol: each model receives a fixed input instance and produces a single prediction, with no web access. 
All retrieval signals must originate solely from the provided document set (Related-only / Full-set) or the oracle concept list (Concepts). 
This design enforces the intended capability decoupling between (i) knowledge selection and de-noising, (ii) document-to-concept extraction, and (iii) concept-level composition and scheduling.

Because some models cannot reliably ingest the full document set (e.g., due to context-length limits that can lead to runtime errors or empty generations), we apply a lightweight pre-processing utility to ensure fair evaluation across models. 

Specifically, we use a deterministic long-text truncation rule that takes an input string \texttt{content} and a maximum character budget \texttt{max\_chars} (default: 30000). 
If \texttt{max\_chars} is \texttt{None}, negative, or the input length is below the limit, the utility returns \texttt{content} unchanged. 
Otherwise, it sets \texttt{half = max\_chars // 2}, keeps the first \texttt{half} characters and the last \texttt{half} characters, and inserts a fixed marker line in between: \texttt{[... Observation truncated due to length ...]}. Newlines are added on both sides of the marker.

To reduce evaluation overhead while keeping decoding conditions consistent across models, we use a single shared sampling configuration: temperature = 1 and nucleus sampling with $p=0.7$ (no additional decoding constraints beyond the API defaults). Each model-setting pair is run twice, and we report the average score over the two runs.

To minimize prompt-induced artifacts, we never instruct the model to quote documents verbatim; instead, we ask for concise, verifiable final answers consistent with the benchmark's answer format. 
All raw generations are logged verbatim. Answers are evaluated by doubao-seed-1-6-251015, using a fixed evaluation prompt provided in Appendix~\ref{app:prompt}.

\begin{table}[h]
\centering
\begin{tabular}{lccc p{4cm}}
\toprule
\textbf{Setting} & \textbf{Problem} & \textbf{Concepts} & \textbf{Documents} & \textbf{Capability} \\
\midrule
Instruction-only & \checkmark & - & - & parametric knowledge \\
Concepts-only & \checkmark & \checkmark (oracle) & - & concept-level composition and scheduling reasoning \\
Related-only & \checkmark & - & \checkmark (related doc) & knowledge selection and reasoning\\
Full-set & \checkmark & - & \checkmark (related+noise doc) & knowledge selection, de-noising and reasoning\\
\bottomrule
\end{tabular}
\caption{Evaluation settings with different input configurations}
\label{tab:settings}
\end{table}

We report the answer-level accuracy score as the primary evaluation metric. A prediction is considered correct if it matches the ground-truth answer according to task-specific normalization rules, including numeric tolerance, symbolic equivalence, or checklist-based matching for structured answers. 

Additionally, we use a simple retrieval loss, defined as the difference between the Concepts-only score and the Full-set score: 
$\text{Retrieval Loss (RLoss)} = Score_{\text{Concepts-only}} - Score_{\text{Full-set}}.$
A larger retrieval loss indicates greater degradation due to retrieval.

\subsection{Results and Analysis}
\label{subsec:results}

We present results in three layers: 
(i) overall accuracy across the four evaluation settings,
(ii) controlled factor analyses that stress specific retrieval--reasoning bottlenecks, and
(iii) qualitative CoT-based error attribution.

\paragraph{Overall performance and setting-wise gaps.}
Table~\ref{tab:four_settings_comparison} reports answer-level accuracy for each model under \settingZero, \settingOne, \settingTwo, and \settingThree.
Across models, we observe a consistent ordering \textit{\settingOne}(avg. 75.39) $>$ \textit{\settingTwo}(avg. 62.89) $>$ \textit{\settingThree}(avg. 51.25), indicating that
(i) providing gold concepts substantially improves outcomes by removing concept-extraction errors,
(ii) concept extraction from documents remains a major source of loss, and
(iii) adding distractor documents further degrades performance through evidence selection failures.
Notably, we also observe a counter-intuitive but recurrent phenomenon where \textit{\settingZero}(avg. 55.89) can match or even exceed \textit{\settingThree}(avg. 51.21) for most models, suggesting that the performance drop is not merely attributable to noise volume but to a shift in the model's reasoning regime once external context is introduced. 

\paragraph{Decomposing loss: retrieval vs.\ reasoning.}
The controlled settings allow us to attribute end-to-end degradation to two interpretable gaps:
\emph{document-to-concept loss} (the performance drop from \textit{\settingOne} to \textit{\settingTwo}) and
\emph{noise-induced loss} (the drop from \textit{\settingTwo} to \textit{\settingThree}).
The first gap captures failures to \emph{recover and use} the required concepts from evidence---including locating the right passages, extracting the relevant conceptual content, and operationalizing it as executable reasoning steps.
The second gap isolates robustness to distractors: given that the necessary evidence is present, it measures whether the model can ignore topic-adjacent noise and sustain a correct, evidence-conditioned reasoning trajectory.
Overall, this decomposition matches the benchmark’s intended separation of (i) evidence selection / de-noising, (ii) structured concept extraction from long-context documents, and (iii) concept-level coordination and scheduling in multi-step reasoning.

\begin{figure}[h]
    \centering
    \includegraphics[width=.8\textwidth]{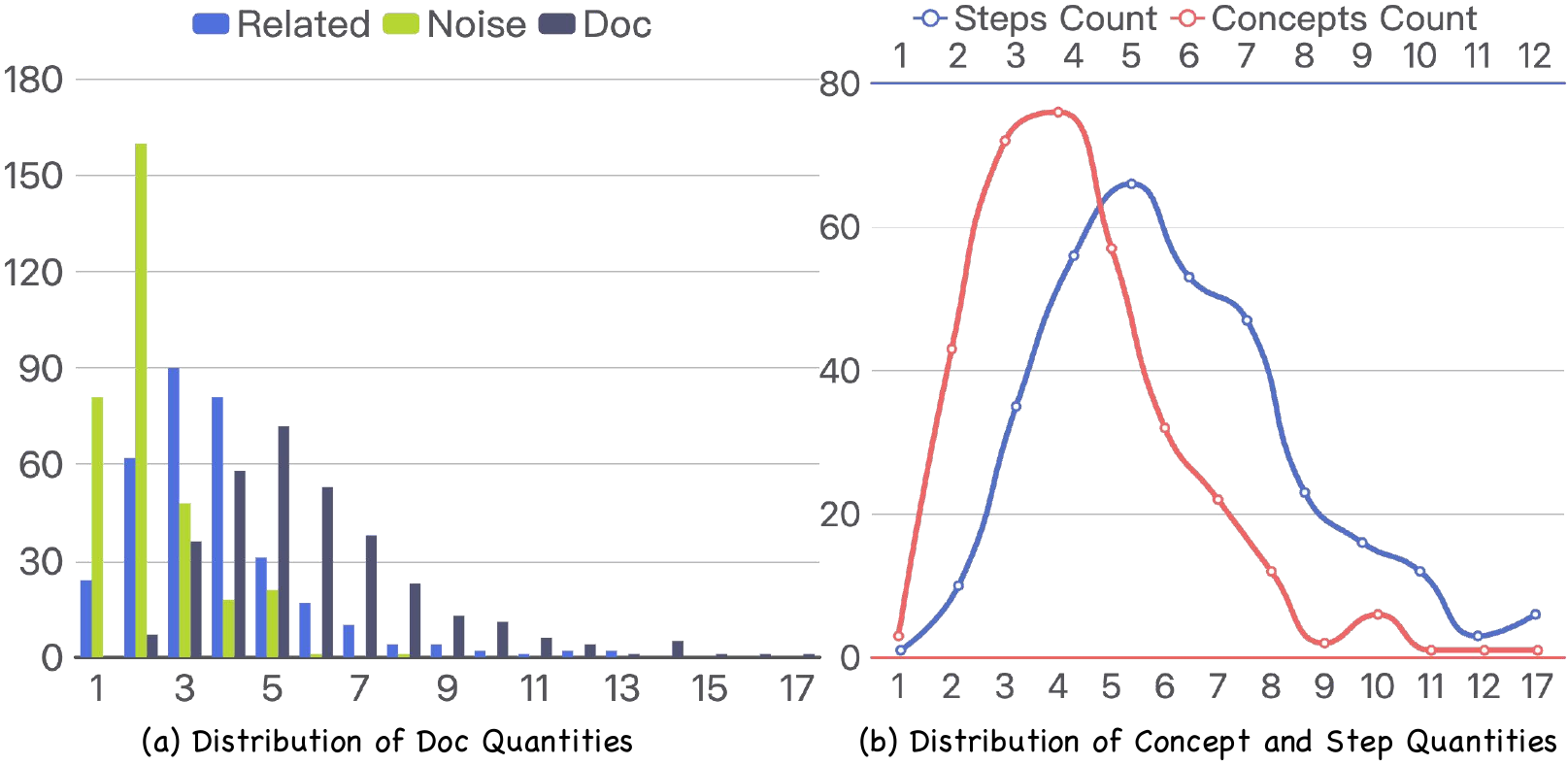}
    \caption{Structural and contextual complexity of the dataset. 
\textbf{(a)} Reports the distributions of reasoning step counts and required concept counts per problem, reflecting multi-step and multi-concept reasoning demands. 
\textbf{(b)} Presents the distributions of related documents, noise documents, and total documents per instance, characterizing the degree of retrieval difficulty and noise exposure in the document sets. 
Together, these statistics illustrate the controlled diversity of DeR² across reasoning depth, conceptual load, and retrieval conditions.
}
    \label{fig:line}
\end{figure}

\paragraph{Controlled factor analysis.}
We next stratify instances by dataset factors that are designed to surface distinct failure modes.

\begin{figure}[h]
    \centering
    \includegraphics[width=.9\textwidth]{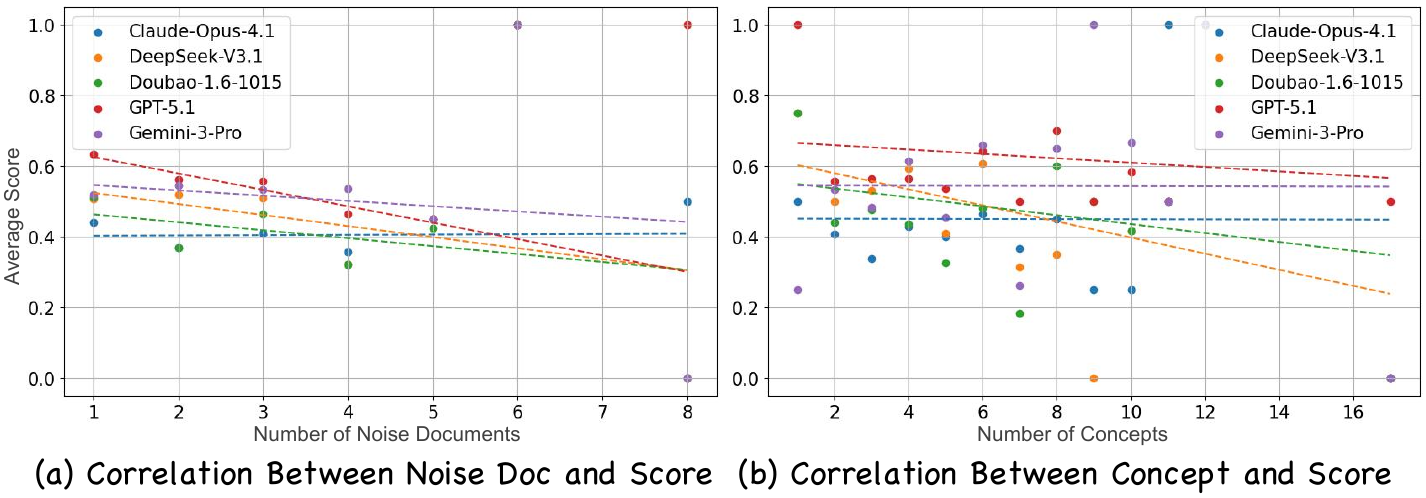}
    \caption{\textbf{Impact of Document Characteristics on Model Performance.}
    (a) Under the \textit{Full-set} setting, model score as a function of the number of noise documents.
    (b) Under the \textit{Full-set} setting, model score as a function of the number of concepts.}
    \label{fig:doccount2score}
\end{figure}

\textbf{Noise documents.}
As the number of noise documents increases, \textit{Full-set} performance exhibits a gradual but non-linear decline, consistent with the hypothesis that distractors do not simply 'dilute' signal, but can redirect early-step abstraction and cause irreversible trajectory drift, as illustrated in Fig.\ref{fig:doccount2score}.
We additionally compare cases where models retrieve relevant concepts but fail to \emph{use} them: 
A frequent pattern is that the model cites superficially relevant evidence while executing a generic template solution, indicating evidence is present but not integrated into the causal chain of inference.

\textbf{Required concept count.}
Accuracy decreases as the number of required concepts grows, even under the \textit{\settingOne} setting.
This shows that providing all necessary concepts does not eliminate errors: models still struggle with concept-level working memory, dependency tracking, and coordinating multiple constraints.
The effect is amplified in \textit{\settingOne} and \textit{\settingThree}, where concept extraction errors and spurious concept injection further enlarge the hypothesis space and increase coordination burden. 

\textbf{Reasoning depth and answer type.}
We bucket problems by reasoning depth (e.g., number of steps in the reference CoT) and by answer type (e.g., formula derivation, numeric, boolean, or conclusion-style). 
Deeper reasoning typically yields larger gaps between \textit{\settingOne} and document-based settings, reflecting compounding failures in multi-step grounding and intermediate-state maintenance. 
In parallel, formula-derivation items benefit the most from explicit concepts, suggesting that document-based settings often fail at operationalizing constructive or procedural concepts (e.g., algorithm instantiation, theorem application) even when high-level definitions are recognized. 

\paragraph{Chain-of-thought and error attribution.}
To move beyond accuracy, we analyze predicted CoTs against reference concepts and reasoning structure.
We categorize failures into three dominant modes:

\textbf{(1) Reasoning mode switching failure.}
When external documents are provided, models often abandon a viable parametric reasoning path but fail to reliably anchor a new evidence-driven chain.
This produces the paradoxical \textit{\settingZero} $\ge$ \textit{\settingThree} behavior on a non-trivial subset (e.g., Gemini-3-Pro-Preview scores 64.2 under \settingZero, but only 53.7 when provided with more knowledge in the \settingThree; Claude-Opus-4.1-thinking scores 49.3 under \settingZero, but only 40 in the \settingThree; similar patterns are observed for models like OpenAI-GPT-5.1-high, DeepSeek-V3.2-Exp-Thinking, etc.) and indicates that \emph{switch control} between internal-knowledge reasoning and context-grounded reasoning is itself a bottleneck.

\textbf{(2) Structural retrieval errors.}
Errors frequently arise not from missing definitions but from failing to execute constructive mechanisms described in documents (e.g., algorithmic steps, instantiation details, auxiliary-variable constructions).
Models \emph{replace these mechanisms with familiar heuristics}, yielding CoTs that are locally coherent but globally incorrect. 

\textbf{(3) Concept coordination breakdown.}
Even with correct concepts available (in \textit{Concepts}), models sometimes fail due to poor coordination:
They activate only a subset of necessary constraints, apply concepts out of order, or lose track of
intermediate invariants. This highlights that retrieval-infused reasoning is not a simple
'retrieval + reasoning' composition; it requires \emph{explicit scheduling and multi-constraint control}. 

\begin{table}[t]
\centering
\small
\begin{tabular}{lccccc}
\toprule
\textbf{Model} 
& \textbf{\settingZero} 
& \textbf{\settingThree} 
& \textbf{\settingTwo} 
& \textbf{\settingOne} 
& \textbf{RLoss} \\
\midrule
OpenAI-GPT-5.2-high                 & \textbf{65.8}	& \textbf{71.1}	& \textbf{71.4}	& \textbf{83.8} & \underline{12.7}  \\
Gemini-3-Pro-Preview                & \underline{64.2}	& 53.7	& \underline{68.3}	& 80.9	& 27.2  \\
Gemini-3-Flash-Preview              & 60.5	& \underline{66}.0	& 64.8	& 78.3	& \textbf{12.3} \\
OpenAI-GPT-5.1-high                 & 59.8	& 57.0	& 66.9	& \underline{81.4}	& 24.4 \\
DeepSeek-V3.2-Exp-Thinking          & 57.6	& 49.3	& 61.3	& 75.3	& 26.0 \\
Moonshot-kimi-k2-thinking           & 55.9	& 52.7	& 65.0	& 74.4	& 21.7 \\
Gemini-2.5-Pro                      & 54.0	& 51.5	& 66.6	& 78.5	& 27.0 \\
GLM-4-6                             & 53.9	& 43.2	& 57.7	& 70.8	& 27.6 \\
QwenAPI-3-max-0923                  & 53.2	& 41.8	& 61.1	& 70.6	& 28.8 \\
Claude-Sonnet-4.5                   & 53.2	& 45.2	& 59.1	& 74.1	& 28.9 \\
Doubao-1.8-1228-high                & 52.3	& 52.0	& 64.8	& 74.3	& 22.3 \\
DeepSeek-V3.1-terminus-thinking     & 52.2	& 50.2	& 62.1	& 69.8	& 19.6 \\
Doubao-1.6-1015-high                & 50.6	& 43.3	& 59.4	& 70.8	& 27.5 \\
Claude-Opus-4.1-thinking            & 49.3	& 40.0	& 52.0	& 72.4	& 32.4 \\
\midrule
Average Score                       & 55.9 & 51.2 & 62.9 & 75.4 & 24.2 \\
\bottomrule
\end{tabular}
\caption{Answer-level accuracy (\%) under four controlled evaluation settings, ranked by RE score. \textbf{Bold} indicates best performance, \underline{underline} indicates second-best per column. Instruction-only reflects parametric knowledge; Concepts-only provides an empirical upper bound on concept-level reasoning; Related-only and Full-set evaluate retrieval-conditioned reasoning under weak and noisy evidence, respectively.}
\label{tab:four_settings_comparison}
\end{table}

Table \ref{tab:error_attribution} presents a detailed breakdown of the main failure modes at the reasoning chain level, covering four controlled evaluation settings. Each sub-table within the main table includes two key types of information: first, the concept-level performance metrics (precision, recall, and accuracy) calculated via a model-assisted annotation protocol; second, the distribution of error types extracted from the predicted reasoning trajectories of evaluated models.

For the concept-level metrics, the calculation process follows a strict workflow: we first use the model \texttt{doubao-seed-1.6-251015} to extract the set of concepts explicitly applied by each evaluated model from its predicted chain-of-thought (\texttt{pred\_cot}). Then, with the corresponding ground-truth concept set (\texttt{true\_concepts}) as a reference, the same \texttt{doubao-seed-1.6-251015} model is prompted to align the predicted and reference concepts, and output the counts of true positives (TP), false positives (FP), and false negatives (FN). Precision and recall are computed using standard formulas based on these counts, while concept accuracy is derived by aggregating the per-instance performance results across the entire dataset.

For the error type distribution, the data is collected through a targeted sampling and analysis process: for each evaluated model and each of the four evaluation settings, we uniformly sample 50 instances where the model’s final answer is incorrect (to focus on failure modes beyond simple concept omission). We then use \texttt{doubao-seed-1.6-251015} as a structured evaluator, providing it with four inputs for each sampled instance—the predicted concepts (\texttt{pred\_concept}), predicted chain-of-thought (\texttt{pred\_cot}), ground-truth concepts (\texttt{true\_concept}), and reference reasoning (\texttt{true\_cot})—and prompt it to diagnose the root causes of failure. Each incorrect instance is labeled with one or more of four predefined error categories: \textbf{(1) reasoning-process error} (flaws in the logical flow of reasoning), \textbf{(2) numeric or formalization error} (mistakes in calculations or symbolic representation), \textbf{(3) missing core concept} (failure to include a key concept required for the correct answer), and \textbf{(4) incorrect understanding or misuse of a core concept} (misinterpreting a concept or applying it in an inappropriate context). The full prompt used to guide the \texttt{doubao-seed-1.6-251015} evaluator during the error diagnosis process is provided in the appendix.\ref{app:prompt} for reproducibility.

In \textit{\settingZero}, the errors mainly manifest as missing core concepts and issues with general reasoning heuristics, reflecting the limitations of pure parametric knowledge. By \textit{\settingOne}, overall accuracy significantly improves, and errors related to concepts are largely eliminated; the remaining failures are predominantly procedural, indicating that simply retrieving the correct concepts does not guarantee the correct execution of multi-step reasoning.

It is further observed that in the \settingThree, \settingTwo, and \settingZero settings, the missing core concept(MC) is the most frequent error. This aligns with the knowledge input scaling relationship, but it is worth noting that the gap in the number of missing core concepts between \settingZero and the knowledge-providing settings (\settingThree and \settingTwo) is not large, with all hovering around 40, reflecting a limitation in the model’s ability to retrieve the correct knowledge points, or the model sometimes does not even know which concepts to extract to answer the current question. Additionally, Gemini-3-Pro only made 29 errors in the \settingTwo setting and delivered its optimal performance in \settingZero (with merely 34 mistakes). This indicates that Gemini may possess more endogenous knowledge, enabling it to retrieve and comprehend concepts effectively even with limited background information.

In \textit{\settingOne}, the model must recover concepts from less relevant evidence, and both concept omission and reasoning errors reappear, revealing structural retrieval failure rather than simple definition gaps. Finally, in the \textit{\settingThree} setting, although some retrievals are successful, the presence of topic-related noise exacerbates reasoning errors and misalignment of concepts. 

Overall, these tables indicate that the performance degradation under retrieval-augmented settings is not only due to missing concepts but also stems from failures in procedural and conceptual coordination during evidence-conditioned reasoning.

\begin{table*}[t]
\centering
\small
\begin{tabular}{lcccc}
\toprule
\textbf{Model} & \textbf{Prec.} & \textbf{Rec.} & \textbf{Score} & \textbf{Err.-type dist.} \\
\midrule
\multicolumn{5}{l}{\textit{\textbf{\settingThree setting}}} \\
\midrule
Claude-Opus-4.1     & 0.242 & 0.224 & 40.0 & MC:42 UC:7 \textbf{R:35} \underline{NF:3} \\
DeepSeek-V3.1       & \underline{0.323} & 0.283 & 50.2 & MC:42 \underline{UC:5} R:44 \textbf{NF:2} \\
Doubao-1.6-1015     & 0.270 & 0.237 & 43.3 & \textbf{MC:37} UC:10 R:43 NF:4 \\
Gemini-3-Pro        & \textbf{0.333} & \underline{0.315} & \underline{53.7} & MC:41 UC:9 \underline{R:38} NF:8 \\
GPT-5.1             & 0.318 & \textbf{0.330} & \textbf{57.0} & \underline{MC:39} \textbf{UC:4} R:41 \underline{NF:3} \\
\textit{Average}             & 0.297 & 0.278 & 48.8 & MC:40.2 UC:7 R:40.2 NF:4 \\
\midrule
\multicolumn{5}{l}{\textit{\textbf{Related-only setting}}} \\
\midrule
Claude-Opus-4.1     & 0.241 & 0.212 & 52.0 & MC:38 UC:7 R:38 \underline{NF:2} \\
DeepSeek-V3.1       & 0.308 & 0.270 & 62.1 & MC:37 UC:4 R:40 \underline{NF:2} \\
Doubao-1.6-1015     & 0.276 & 0.247 & 59.4 & \underline{MC:32} UC:6 \underline{R:37} \underline{NF:2} \\
Gemini-3-Pro        & \textbf{0.349} & \underline{0.332} & \textbf{68.3} & \textbf{MC:29} \underline{UC:2} \underline{R:37} NF:3 \\
GPT-5.1             & \underline{0.329} & \textbf{0.339} & \underline{66.9} & MC:34 \textbf{UC:1} \textbf{R:33} NF:5 \\
\textit{Average}             & 0.301 & 0.280 & 61.7 & MC:34 UC:4 R:37 NF:2.8 \\
\midrule
\multicolumn{5}{l}{\textit{\textbf{Concepts-only setting}}} \\
\midrule
Claude-Opus-4.1     & 0.718 & 0.651 & 72.4 & MC:9 UC:5 R:11 \underline{NF:0} \\
DeepSeek-V3.1       & 0.695 & 0.650 & 69.8 & MC:13 UC:6 \underline{R:7} NF:2 \\
Doubao-1.6-1015     & 0.724 & 0.648 & 70.8 & MC:11 UC:5 R:10 \underline{NF:0} \\
Gemini-3-Pro        & \underline{0.740} & \underline{0.730} & \textbf{80.9} & \underline{MC:7} \underline{UC:2} \underline{R:7} \underline{NF:0} \\
GPT-5.1             & \textbf{0.754} & \textbf{0.754} & \underline{81.4} & \textbf{MC:6} \textbf{UC:0} \underline{R:7} \underline{NF:0} \\
\textit{Average}             & 0.766 & 0.686 & 75.0 & MC:9.2 UC:3.8 R:8.4 NF:0.4 \\
\midrule
\multicolumn{5}{l}{\textit{\textbf{Instruction-only setting}}} \\
\midrule
Claude-Opus-4.1     & 0.246 & 0.236 & 49.3 & MC:42 UC:4 \textbf{R:30} \underline{NF:3} \\
DeepSeek-V3.1       & \underline{0.301} & 0.260 & 52.2 & MC:40 UC:4 R:43 NF:4 \\
Doubao-1.6-1015     & 0.293 & 0.267 & 50.6 & \underline{MC:39} UC:3 R:41 NF:4 \\
Gemini-3-Pro        & \textbf{0.341} & \textbf{0.336} & \textbf{64.2} & \textbf{MC:34} \textbf{UC:0} R:42 \textbf{NF:0} \\
GPT-5.1             & 0.300 & \underline{0.297} & \underline{59.8} & MC:42 \underline{UC:2} \underline{R:38} NF:5 \\
\textit{Average}             & 0.296 & 0.279 & 55.2 & MC:39.4 UC:2.6 R:38.8 NF:3.2 \\
\bottomrule
\end{tabular}
\caption{CoT-based error attribution under four evaluation settings. Abbreviations: MC=missing core concept; UC=misused/incorrect core concept; R=reasoning-process error; NF=numeric/formalization error.}
\label{tab:error_attribution}
\end{table*}

Additionally, in the concepts-only setting, the model’s precision and recall both fail to reach 1, indicating that the model cannot comprehend the provided unseen concepts, even when usage instructions for those concepts are supplied.

\section{Related Work}

\subsection{Existing Benchmarks}
Early reading comprehension benchmarks such as SQuAD \cite{rajpurkar2016squad} and Natural Questions \cite{kwiatkowski2019natural} established evaluation protocols for extractive question answering over short passages. Multi-hop extensions such as HotpotQA \cite{DBLP:conf/emnlp/Yang0ZBCSM18} increased reasoning complexity by requiring evidence aggregation across passages, though answers remained primarily extractive. While recent variants incorporate more abstractive generation, evaluation metrics, and task designs still tend to favor localized extraction over genuine multi-document synthesis \cite{kamalloo2023limitations}.

Retrieval-augmented generation benchmarks, including KILT \cite{petroni2021kilt}, and FreshQA \cite{Vu2023FreshLLMsRL}, evaluate end-to-end RAG pipelines over large corpora. By coupling retrieval with generation, they measure system-level performance, but this coupling conflates multiple capabilities, making error attribution difficult: failures may stem from retrieval gaps, reasoning deficits, or evidence integration issues. Moreover, many tasks emphasize factoid lookup or temporal fact verification rather than multi-step synthesis that requires integrating novel concepts across multiple documents.

In parallel, mathematical and formal reasoning benchmarks evaluate deductive capabilities largely in isolation from document comprehension. Benchmarks like GSM8K \cite{cobbe2021training}, MATH \cite{hendrycks2021math}, Omni-MATH \cite{gao2024omni} and Putnam-AXIOM \cite{fronsdal2024putnamaxiom} and test multi-step problem solving with increasing complexity. Graduate-level benchmarks such as GPQA \cite{rein2023gpqa} and broad suites like BIG-Bench \cite{DBLP:journals/corr/abs-2206-04615} increase difficulty but remain predominantly parametric: models rely on training-time knowledge rather than synthesizing information from provided evidence.

Finally, long-context evaluations (e.g., recent long-context model work \cite{geminiteam2024gemini15unlockingmultimodal}) demonstrate improved capacity for processing extensive documents, yet they still lack systematic measurement of synthesis when evidence must be extracted from noisy, realistic corpora. Related work on document-level reasoning \cite{kamalloo2023limitations, zhao2023docmath} highlights limitations of current QA benchmarks, but does not provide controlled environments for measuring multi-document synthesis under adversarial distractors.

\subsection{ Decoupled Retrieval and Reasoning Benchmark}
Our benchmark is designed as a controlled environment that disentangles retrieval from reasoning for document-grounded scientific synthesis. Concretely, each instance is organized around a five-field structure (Instruction, Concepts, CoT, Answer, Document Set) and a two-stage verifiability protocol: problems are unsolvable in instruction-only mode, yet become solvable once the correct concepts are provided. This explicitly targets scientific reasoning that requires integrating novel concepts rather than relying on parametric recall.

Unlike end-to-end RAG benchmarks that conflate retrieval and generation, we introduce evaluation settings that decouple capabilities by construction: \textbf{Instruction-only} isolates parametric knowledge; \textbf{Concepts} isolates concept-level reasoning; \textbf{Related-only} measures weak concept retrieval plus reasoning; and \textbf{Full-set} stress-tests evidence selection and de-noising under distractors. This structure enables attribution of performance loss to (i) evidence selection and de-noising, (ii) document-to-concept extraction, and (iii) concept-level composition and scheduling.

Finally, by explicitly including topically adjacent but concept-irrelevant distractors in the document set, our benchmark operationalizes adversarial noise as a first-class variable. This targets failure modes that are typically masked in prior benchmarks—especially the difficulty of switching from parametric reasoning to evidence-grounded reasoning, and the brittleness of multi-concept coordination under evidence overload.

\section{Conclusion}
\label{sec:conclusion}

We introduce \textsc{DeR$^{2}$}, a controlled benchmark for \emph{decoupling} retrieval from reasoning in retrieval-infused problem solving. Each instance is built with a two-stage verifiability protocol and a structured doc set (related + distractors), enabling stable attribution of performance loss to (i) evidence selection/de-noising, (ii) document-to-concept extraction, and (iii) concept-level composition and scheduling. 

Across strong commercial and open models, we observe consistent non-trivial failure modes: (1) \textbf{reasoning mode switching failures} where \textbf{Instruction-only can outperform Full-set}, indicating that adding documents may disrupt a previously viable parametric reasoning path; (2) \textbf{structural (procedural) retrieval failures} where models recognize definitions but cannot execute constructive mechanisms; (3) \textbf{Concepts-only is not an upper bound}, revealing bottlenecks in multi-concept coordination and long-range dependency tracking; and (4) \textbf{nonlinear noise effects} that alter early reasoning topology and lead to irrecoverable trajectory drift. 

These results suggest that ``deep research'' capability is \emph{not} a simple extension of end-to-end RAG accuracy: robust systems must explicitly support evidence-grounded mode control, executable concept utilization, and adaptive hypothesis management under noisy contexts. We release \textsc{DeR$^{2}$} to facilitate principled model selection and diagnosis for retrieval-infused reasoning and to provide a practical target for future training and evaluation.

\clearpage

\section*{Contributions and Acknowledgements}

Multimodal Art Projection (M-A-P) is a non-profit open-source AI research community, ran by donation.
The community members are working on research topics in a wide range of spectrum, including but not limited to the pre-training paradigm of foundation models, large-scale data collection and processing, and the derived applications on coding, reasoning and music generation.

\textbf{Leading Authors}
\begin{itemize}
    \item Shuangshuang Ying, M-A-P
    \item Zheyu Wang, ByteDance Seed, China 
    \item Yunjian Peng, ByteDance Seed, China 
    \item Jin Chen, ByteDance Seed, China 
\end{itemize}

\textbf{Contributors}
\begin{itemize}
    \item Yuhao Wu, ByteDance Seed, China 
    \item Hongbin Lin, ByteDance Seed, China 
    \item Dingyu He, ByteDance Seed, China 
    \item Siyi Liu, ByteDance Seed, China 
    \item Gengchen Yu, ByteDance Seed, China 
    \item YinZhu Piao, ByteDance Seed, China 
    \item Yuchen Wu, ByteDance Seed, China 
    \item Xin Gui, M-A-P 
    \item Zhongyuan Peng, Fudan University 
    \item Xin Li, Nanyang Technological University 
    \item Xeron Du, M-A-P 
\end{itemize}

\textbf{Advisors}
    \begin{itemize}
        \item Libo Qin, Harbin Institute of Technology, Shenzhen
        \item YiXin Cao, Fudan University
        \item Stephen Huang, Peking University
    \end{itemize}

\textbf{Corresponding Authors}
    \begin{itemize}
        \item Ge Zhang, ByteDance Seed, China
    \end{itemize}

\clearpage

\clearpage

\bibliographystyle{plainnat}
\bibliography{main.bib}

@article{gao2024omni,
  title={Omni-math: A universal olympiad level mathematic benchmark for large language models},
  author={Gao, Bofei and Song, Feifan and Yang, Zhe and Cai, Zefan and Miao, Yibo and Dong, Qingxiu and Li, Lei and Ma, Chenghao and Chen, Liang and Xu, Runxin and others},
  journal={arXiv preprint arXiv:2410.07985},
  year={2024}
}

@misc{claude4.5,
 author = {anthropic},
 title = {Claude 4.5 Sonnet},
 URL = {https://www.anthropic.com/news/claude-sonnet-4-5},
 year = {2025}
}

@misc{zhipu2025glm46,
  title={GLM-4.6: An Open Large Language Model},
  author={Zhipu AI},
  year={2025},
  url={https://huggingface.co/zai-org/GLM-4.6},
  note={Hugging Face Model Card}
}

@misc{bytedance2024doubao,
  title={Doubao Large Language Model},
  author={ByteDance},
  year={2024},
  url={https://www.doubao.com/},
  note={Accessed: 2025-01-16}
}

@article{deepseek2024v32,
  title={DeepSeek-V3.2: Pushing the Frontier of Open Large Language Models},
  author={DeepSeek-AI},
  journal={arXiv preprint arXiv:2512.02556},
  year={2024},
  url={https://arxiv.org/abs/2512.02556}
}

@article{yang2025qwen3,
  title={Qwen3 technical report},
  author={Yang, An and Li, Anfeng and Yang, Baosong and Zhang, Beichen and Hui, Binyuan and Zheng, Bo and Yu, Bowen and Gao, Chang and Huang, Chengen and Lv, Chenxu and others},
  journal={arXiv preprint arXiv:2505.09388},
  year={2025}
}

@misc{claude-opus-4-1,
 author = {anthropic},
 title = {claude-opus-4-1},
 URL = {https://www.anthropic.com/news/claude-opus-4-1
},
 year = {2025}
}

@misc{openai2025gpt51,
  author = {OpenAI},
  title = {Introducing GPT-5.1}, 
  year = {2025}, 
  url = {https://openai.com/index/gpt-5-1/}
}

@misc{kimiteam2025kimik2openagentic,
      title={Kimi K2: Open Agentic Intelligence}, 
      author={Kimi Team and Yifan Bai and Yiping Bao and Guanduo Chen and Jiahao Chen and Ningxin Chen and Ruijue Chen and Yanru Chen and Yuankun Chen and Yutian Chen and Zhuofu Chen and Jialei Cui and Hao Ding and Mengnan Dong and Angang Du and Chenzhuang Du and Dikang Du and Yulun Du and Yu Fan and Yichen Feng and Kelin Fu and Bofei Gao and Hongcheng Gao and Peizhong Gao and Tong Gao and Xinran Gu and Longyu Guan and Haiqing Guo and Jianhang Guo and Hao Hu and Xiaoru Hao and Tianhong He and Weiran He and Wenyang He and Chao Hong and Yangyang Hu and Zhenxing Hu and Weixiao Huang and Zhiqi Huang and Zihao Huang and Tao Jiang and Zhejun Jiang and Xinyi Jin and Yongsheng Kang and Guokun Lai and Cheng Li and Fang Li and Haoyang Li and Ming Li and Wentao Li and Yanhao Li and Yiwei Li and Zhaowei Li and Zheming Li and Hongzhan Lin and Xiaohan Lin and Zongyu Lin and Chengyin Liu and Chenyu Liu and Hongzhang Liu and Jingyuan Liu and Junqi Liu and Liang Liu and Shaowei Liu and T. Y. Liu and Tianwei Liu and Weizhou Liu and Yangyang Liu and Yibo Liu and Yiping Liu and Yue Liu and Zhengying Liu and Enzhe Lu and Lijun Lu and Shengling Ma and Xinyu Ma and Yingwei Ma and Shaoguang Mao and Jie Mei and Xin Men and Yibo Miao and Siyuan Pan and Yebo Peng and Ruoyu Qin and Bowen Qu and Zeyu Shang and Lidong Shi and Shengyuan Shi and Feifan Song and Jianlin Su and Zhengyuan Su and Xinjie Sun and Flood Sung and Heyi Tang and Jiawen Tao and Qifeng Teng and Chensi Wang and Dinglu Wang and Feng Wang and Haiming Wang and Jianzhou Wang and Jiaxing Wang and Jinhong Wang and Shengjie Wang and Shuyi Wang and Yao Wang and Yejie Wang and Yiqin Wang and Yuxin Wang and Yuzhi Wang and Zhaoji Wang and Zhengtao Wang and Zhexu Wang and Chu Wei and Qianqian Wei and Wenhao Wu and Xingzhe Wu and Yuxin Wu and Chenjun Xiao and Xiaotong Xie and Weimin Xiong and Boyu Xu and Jing Xu and Jinjing Xu and L. H. Xu and Lin Xu and Suting Xu and Weixin Xu and Xinran Xu and Yangchuan Xu and Ziyao Xu and Junjie Yan and Yuzi Yan and Xiaofei Yang and Ying Yang and Zhen Yang and Zhilin Yang and Zonghan Yang and Haotian Yao and Xingcheng Yao and Wenjie Ye and Zhuorui Ye and Bohong Yin and Longhui Yu and Enming Yuan and Hongbang Yuan and Mengjie Yuan and Haobing Zhan and Dehao Zhang and Hao Zhang and Wanlu Zhang and Xiaobin Zhang and Yangkun Zhang and Yizhi Zhang and Yongting Zhang and Yu Zhang and Yutao Zhang and Yutong Zhang and Zheng Zhang and Haotian Zhao and Yikai Zhao and Huabin Zheng and Shaojie Zheng and Jianren Zhou and Xinyu Zhou and Zaida Zhou and Zhen Zhu and Weiyu Zhuang and Xinxing Zu},
      year={2025},
      eprint={2507.20534},
      archivePrefix={arXiv},
      primaryClass={cs.LG},
      url={https://arxiv.org/abs/2507.20534}, 
}

@article{kwiatkowski2019natural,
  title={Natural questions: a benchmark for question answering research},
  author={Kwiatkowski, Tom and Palomaki, Jennimaria and Redfield, Olivia and Collins, Michael and Parikh, Ankur and Alberti, Chris and Epstein, Danielle and Polosukhin, Illia and Devlin, Jacob and Lee, Kenton and others},
  journal={Transactions of the Association for Computational Linguistics},
  volume={7},
  pages={453--466},
  year={2019},
  publisher={MIT Press One Rogers Street, Cambridge, MA 02142-1209, USA journals-info~…}
}

@article{rajpurkar2016squad,
  title={Squad: 100,000+ questions for machine comprehension of text},
  author={Rajpurkar, Pranav and Zhang, Jian and Lopyrev, Konstantin and Liang, Percy},
  journal={arXiv preprint arXiv:1606.05250},
  year={2016}
}

@misc{cobbe2021training,
      title={Training Verifiers to Solve Math Word Problems}, 
      author={Karl Cobbe and Vineet Kosaraju and Mohammad Bavarian and Mark Chen and Heewoo Jun and Lukasz Kaiser and Matthias Plappert and Jerry Tworek and Jacob Hilton and Reiichiro Nakano and Christopher Hesse and John Schulman},
      year={2021},
      eprint={2110.14168},
      archivePrefix={arXiv},
      primaryClass={cs.LG}
}

@misc{geminiteam2024gemini15unlockingmultimodal,
      title={Gemini 1.5: Unlocking multimodal understanding across millions of tokens of context}, 
      author={Gemini Team},
      year={2024},
      eprint={2403.05530},
      archivePrefix={arXiv},
      primaryClass={cs.CL},
      url={https://arxiv.org/abs/2403.05530}, 
}

@article{DBLP:journals/corr/abs-2206-04615,
  author       = {Aarohi Srivastava and
                  Abhinav Rastogi and
                  Abhishek Rao and
                  Abu Awal Md Shoeb and
                  Abubakar Abid and
                  Adam Fisch and
                  Adam R. Brown and
                  Adam Santoro and
                  Aditya Gupta and
                  Adri{\`{a}} Garriga{-}Alonso and
                  Agnieszka Kluska and
                  Aitor Lewkowycz and
                  Akshat Agarwal and
                  Alethea Power and
                  Alex Ray and
                  Alex Warstadt and
                  Alexander W. Kocurek and
                  Ali Safaya and
                  Ali Tazarv and
                  Alice Xiang and
                  Alicia Parrish and
                  Allen Nie and
                  Aman Hussain and
                  Amanda Askell and
                  Amanda Dsouza and
                  Ameet Rahane and
                  Anantharaman S. Iyer and
                  Anders Andreassen and
                  Andrea Santilli and
                  Andreas Stuhlm{\"{u}}ller and
                  Andrew M. Dai and
                  Andrew La and
                  Andrew K. Lampinen and
                  Andy Zou and
                  Angela Jiang and
                  Angelica Chen and
                  Anh Vuong and
                  Animesh Gupta and
                  Anna Gottardi and
                  Antonio Norelli and
                  Anu Venkatesh and
                  Arash Gholamidavoodi and
                  Arfa Tabassum and
                  Arul Menezes and
                  Arun Kirubarajan and
                  Asher Mullokandov and
                  Ashish Sabharwal and
                  Austin Herrick and
                  Avia Efrat and
                  Aykut Erdem and
                  Ayla Karakas and
                  et al.},
  title        = {Beyond the Imitation Game: Quantifying and extrapolating the capabilities
                  of language models},
  journal      = {CoRR},
  volume       = {abs/2206.04615},
  year         = {2022},
  url          = {https://doi.org/10.48550/arXiv.2206.04615},
  doi          = {10.48550/arXiv.2206.04615},
  eprinttype    = {arXiv},
  eprint       = {2206.04615},
  bibsource    = {dblp computer science bibliography, https://dblp.org}
}

@article{hendrycks2021math,
  title={Measuring mathematical problem solving with the math dataset},
  author={Hendrycks, Dan and Burns, Collin and Kadavath, Saurav and Arora, Akul and Basart, Steven and Tang, Eric and Song, Dawn and Steinhardt, Jacob},
  journal={arXiv preprint arXiv:2103.03874},
  year={2021}
}

@inproceedings{Vu2023FreshLLMsRL,
  title={FreshLLMs: Refreshing Large Language Models with Search Engine Augmentation},
  author={Tu Vu and Mohit Iyyer and Xuezhi Wang and Noah Constant and Jerry Wei and Jason Wei and Chris Tar and Yun-Hsuan Sung and Denny Zhou and Quoc Le and Thang Luong},
  booktitle={Annual Meeting of the Association for Computational Linguistics},
  year={2023},
  url={https://api.semanticscholar.org/CorpusID:263672149}
}

@article{zhao2023docmath,
  title={DocMath-eval: Evaluating math reasoning capabilities of LLMs in understanding long and specialized documents},
  author={Zhao, Yilun and Long, Yitao and Liu, Hongjun and Kamoi, Ryo and Nan, Linyong and Chen, Lyuhao and Liu, Yixin and Tang, Xiangru and Zhang, Rui and Cohan, Arman},
  journal={arXiv preprint arXiv:2311.09805},
  year={2023}
}

@inproceedings{petroni2021kilt,
    title = "{KILT}: a Benchmark for Knowledge Intensive Language Tasks",
    author = {Petroni, Fabio  and
      Piktus, Aleksandra  and
      Fan, Angela  and
      Lewis, Patrick  and
      Yazdani, Majid  and
      De Cao, Nicola  and
      Thorne, James  and
      Jernite, Yacine  and
      Karpukhin, Vladimir  and
      Maillard, Jean  and
      Plachouras, Vassilis  and
      Rockt{\"a}schel, Tim  and
      Riedel, Sebastian},
    editor = "Toutanova, Kristina  and
      Rumshisky, Anna  and
      Zettlemoyer, Luke  and
      Hakkani-Tur, Dilek  and
      Beltagy, Iz  and
      Bethard, Steven  and
      Cotterell, Ryan  and
      Chakraborty, Tanmoy  and
      Zhou, Yichao",
    booktitle = "Proceedings of the 2021 Conference of the North American Chapter of the Association for Computational Linguistics: Human Language Technologies",
    month = jun,
    year = "2021",
    address = "Online",
    publisher = "Association for Computational Linguistics",
    url = "https://aclanthology.org/2021.naacl-main.200/",
    doi = "10.18653/v1/2021.naacl-main.200",
    pages = "2523--2544"
}

@inproceedings{DBLP:conf/emnlp/Yang0ZBCSM18,
  author       = {Zhilin Yang and
                  Peng Qi and
                  Saizheng Zhang and
                  Yoshua Bengio and
                  William W. Cohen and
                  Ruslan Salakhutdinov and
                  Christopher D. Manning},
  editor       = {Ellen Riloff and
                  David Chiang and
                  Julia Hockenmaier and
                  Jun'ichi Tsujii},
  title        = {HotpotQA: {A} Dataset for Diverse, Explainable Multi-hop Question
                  Answering},
  booktitle    = {Proceedings of the 2018 Conference on Empirical Methods in Natural
                  Language Processing, Brussels, Belgium, October 31 - November 4, 2018},
  pages        = {2369--2380},
  publisher    = {Association for Computational Linguistics},
  year         = {2018},
  url          = {https://doi.org/10.18653/v1/d18-1259},
  doi          = {10.18653/v1/d18-1259},
  bibsource    = {dblp computer science bibliography, https://dblp.org}
}

@misc{google2025gemini3pro,
  title={Gemini 3 Pro: Advanced AI Model},
  author={{Google DeepMind}},
  howpublished={\url{https://deepmind.google/models/gemini/pro/}},
  year={2025},
  month={nov},
  note={Released November 2025}
}

@article{comanici2025gemini,
  title={Gemini 2.5: Pushing the frontier with advanced reasoning, multimodality, long context, and next generation agentic capabilities},
  author={Comanici, Gheorghe and Bieber, Eric and Schaekermann, Mike and Pasupat, Ice and Sachdeva, Noveen and Dhillon, Inderjit and Blistein, Marcel and Ram, Ori and Zhang, Dan and Rosen, Evan and others},
  journal={arXiv preprint arXiv:2507.06261},
  year={2025}
}

@article{rein2023gpqa,
  title={Gpqa: A graduate-level google-proof q\&a benchmark},
  author={Rein, David and Hou, Betty Li and Stickland, Asa Cooper and Petty, Jackson and Pang, Richard Yuanzhe and Dirani, Julien and Michael, Julian and Bowman, Samuel R},
  journal={arXiv preprint arXiv:2311.12022},
  year={2023}
}

@article{phan2025hle,
title={Humanity's Last Exam},
author={Phan, Long and Gatti, Alice and Han, Ziwen and Li, Nathaniel and 
Hu, Josephina and Zhang, Hugh and Shi, Sean and Choi, Michael and 
Agrawal, Anish and Chopra, Arnav and Khoja, Adam and Kim, Ryan and 
Hausenloy, Jason and Zhang, Oliver and Mazeika, Mantas and
Anderson, Daron and Nguyen, Tung and Mahmood, Mobeen and Feng, Fiona and Feng, Steven Y. and Zhao, Haoran and Yu, Michael and Gangal, Varun and Zou, Chelsea and Wang, Zihan and Wang, Jessica P. and Kumar, Pawan and Pokutnyi, Oleksandr and Gerbicz, Robert and Popov, Serguei and Levin, John-Clark and Kazakov, Mstyslav and Schmitt, Johannes and Galgon, Geoff and Sanchez, Alvaro and Lee, Yongki and Yeadon, Will and Sauers, Scott and Roth, Marc and Agu, Chidozie and Riis, Søren and Giska, Fabian and Utpala, Saiteja and Giboney, Zachary and Goshu, Gashaw M. and Xavier, Joan of Arc and Crowson, Sarah-Jane and Naiya, Mohinder Maheshbhai and Burns, Noah and Finke, Lennart and Cheng, Zerui and Park, Hyunwoo and Fournier-Facio, Francesco and Wydallis, John and Nandor, Mark and Singh, Ankit and Gehrunger, Tim and Cai, Jiaqi and McCarty, Ben and Duclosel, Darling and Nam, Jungbae and Zampese, Jennifer and Hoerr, Ryan G. and Bacho, Aras and Loume, Gautier Abou and Galal, Abdallah and Cao, Hangrui and Garretson, Alexis C and Sileo, Damien and Ren, Qiuyu and Cojoc, Doru and Arkhipov, Pavel and Qazi, Usman and Li, Lianghui and Motwani, Sumeet and Witt, Christian Schroeder de and Taylor, Edwin and Veith, Johannes and Singer, Eric and Hartman, Taylor D. and Rissone, Paolo and Jin, Jaehyeok and Shi, Jack Wei Lun and Willcocks, Chris G. and Robinson, Joshua and Mikov, Aleksandar and Prabhu, Ameya and Tang, Longke and Alapont, Xavier and Uro, Justine Leon and Zhou, Kevin and Santos, Emily de Oliveira and Maksimov, Andrey Pupasov and Vendrow, Edward and Zenitani, Kengo and Guillod, Julien and Li, Yuqi and Vendrow, Joshua and Kuchkin, Vladyslav and Ze-An, Ng and Marion, Pierre and Efremov, Denis and Lynch, Jayson and Liang, Kaiqu and Gritsevskiy, Andrew and Martinez, Dakotah and Pageler, Ben and Crispino, Nick and Zvonkine, Dimitri and Fraga, Natanael Wildner and Soori, Saeed and Press, Ori and Tang, Henry and Salazar, Julian and Green, Sean R. and Brüssel, Lina and Twayana, Moon and Dieuleveut, Aymeric and Rogers, T. Ryan and Zhang, Wenjin and Li, Bikun and Yang, Jinzhou and Rao, Arun and Loiseau, Gabriel and Kalinin, Mikhail and Lukas, Marco and Manolescu, Ciprian and Mishra, Subrata and Kamdoum, Ariel Ghislain Kemogne and Kreiman, Tobias and Hogg, Tad and Jin, Alvin and Bosio, Carlo and Sun, Gongbo and Coppola, Brian P and Tarver, Tim and Heidinger, Haline and Sayous, Rafael and Ivanov, Stefan and Cavanagh, Joseph M and Shen, Jiawei and Imperial, Joseph Marvin and Schwaller, Philippe and Senthilkuma, Shaipranesh and Bran, Andres M and Dehghan, Ali and Algaba, Andres and Verbeken, Brecht and Noever, David and V, Ragavendran P and Schut, Lisa and Sucholutsky, Ilia and Zheltonozhskii, Evgenii and Lim, Derek and Stanley, Richard and Sivarajan, Shankar and Yang, Tong and Maar, John and Wykowski, Julian and Oller, Martí and Sandlin, Jennifer and Sahu, Anmol and Hu, Yuzheng and Fish, Sara and Heydari, Nasser and Apronti, Archimedes and Rawal, Kaivalya and Vilchis, Tobias Garcia and Zu, Yuexuan and Lackner, Martin and Koppel, James and Nguyen, Jeremy and Antonenko, Daniil S. and Chern, Steffi and Zhao, Bingchen and Arsene, Pierrot and Goldfarb, Alan and Ivanov, Sergey and Poświata, Rafał and Wang, Chenguang and Li, Daofeng and Crisostomi, Donato and Achilleos, Andrea and Myklebust, Benjamin and Sen, Archan and Perrella, David and Kaparov, Nurdin and Inlow, Mark H and Zang, Allen and Thornley, Elliott and Orel, Daniil and Poritski, Vladislav and Ben-David, Shalev and Berger, Zachary and Whitfill, Parker and Foster, Michael and Munro, Daniel and Ho, Linh and Hava, Dan Bar and Kuchkin, Aleksey and Lauff, Robert and Holmes, David and Sommerhage, Frank and Schneider, Keith and Kazibwe, Zakayo and Stambaugh, Nate and Singh, Mukhwinder and Magoulas, Ilias and Clarke, Don and Kim, Dae Hyun and Dias, Felipe Meneguitti and Elser, Veit and Agarwal, Kanu Priya and Vilchis, Victor Efren Guadarrama and Klose, Immo and Demian, Christoph and Anantheswaran, Ujjwala and Zweiger, Adam and Albani, Guglielmo and Li, Jeffery and Daans, Nicolas and Radionov, Maksim and Rozhoň, Václav and Ma, Ziqiao and Stump, Christian and Berkani, Mohammed and Platnick, Jacob and Nevirkovets, Volodymyr and Basler, Luke and Piccardo, Marco and Jeanplong, Ferenc and Cohen, Niv and Tkadlec, Josef and Rosu, Paul and Padlewski, Piotr and Barzowski, Stanislaw and Montgomery, Kyle and Menezes, Aline and Patel, Arkil and Wang, Zixuan and Tucker-Foltz, Jamie and Stade, Jack and Goertzen, Tom and Kazemi, Fereshteh and Milbauer, Jeremiah and Ambay, John Arnold and Shukla, Abhishek and Labrador, Yan Carlos Leyva and Givré, Alan and Wolff, Hew and Rossbach, Vivien and Aziz, Muhammad Fayez and Kaddar, Younesse and Chen, Yanxu and Zhang, Robin and Pan, Jiayi and Terpin, Antonio and Muennighoff, Niklas and Schoelkopf, Hailey and Zheng, Eric and Carmi, Avishy and Jones, Adam and Shah, Jainam and Brown, Ethan D. L. and Zhu, Kelin and Bartolo, Max and Wheeler, Richard and Ho, Andrew and Barkan, Shaul and Wang, Jiaqi and Stehberger, Martin and Kretov, Egor and Sridhar, Kaustubh and EL-Wasif, Zienab and Zhang, Anji and Pyda, Daniel and Tam, Joanna and Cunningham, David M. and Goryachev, Vladimir and Patramanis, Demosthenes and Krause, Michael and Redenti, Andrew and Bugas, Daniel and Aldous, David and Lai, Jesyin and Coleman, Shannon and Bahaloo, Mohsen and Xu, Jiangnan and Lee, Sangwon and Zhao, Sandy and Tang, Ning and Cohen, Michael K. and Carroll, Micah and Paradise, Orr and Kirchner, Jan Hendrik and Steinerberger, Stefan and Ovchynnikov, Maksym and Matos, Jason O. and Shenoy, Adithya and Junior, Benedito Alves de Oliveira and Wang, Michael and Nie, Yuzhou and Giordano, Paolo and Petersen, Philipp and Sztyber-Betley, Anna and Shukla, Priti and Crozier, Jonathan and Pinto, Antonella and Verma, Shreyas and Joshi, Prashant and Yong, Zheng-Xin and Tee, Allison and Andréoletti, Jérémy and Weller, Orion and Singhal, Raghav and Zhang, Gang and Ivanov, Alexander and Khoury, Seri and Mostaghimi, Hamid and Thaman, Kunvar and Chen, Qijia and Khánh, Trần Quốc and Loader, Jacob and Cavalleri, Stefano and Szlyk, Hannah and Brown, Zachary and Roberts, Jonathan and Alley, William and Sun, Kunyang and Stendall, Ryan and Lamparth, Max and Reuel, Anka and Wang, Ting and Xu, Hanmeng and Raparthi, Sreenivas Goud and Hernández-Cámara, Pablo and Martin, Freddie and Malishev, Dmitry and Preu, Thomas and Korbak, Tomek and Abramovitch, Marcus and Williamson, Dominic and Chen, Ziye and Bálint, Biró and Bari, M Saiful and Kassani, Peyman and Wang, Zihao and Ansarinejad, Behzad and Goswami, Laxman Prasad and Sun, Yewen and Elgnainy, Hossam and Tordera, Daniel and Balabanian, George and Anderson, Earth and Kvistad, Lynna and Moyano, Alejandro José and Maheshwari, Rajat and Sakor, Ahmad and Eron, Murat and McAlister, Isaac C. and Gimenez, Javier and Enyekwe, Innocent and D.O., Andrew Favre and Shah, Shailesh and Zhou, Xiaoxiang and Kamalov, Firuz and Clark, Ronald and Abdoli, Sherwin and Santens, Tim and Meer, Khalida and Wang, Harrison K and Ramakrishnan, Kalyan and Chen, Evan and Tomasiello, Alessandro and Luca, G. Bruno De and Looi, Shi-Zhuo and Le, Vinh-Kha and Kolt, Noam and Mündler, Niels and Semler, Avi and Rodman, Emma and Drori, Jacob and Fossum, Carl J and Jagota, Milind and Pradeep, Ronak and Fan, Honglu and Shah, Tej and Eicher, Jonathan and Chen, Michael and Thaman, Kushal and Merrill, William and Harris, Carter and Gross, Jason and Gusev, Ilya and Sharma, Asankhaya and Agnihotri, Shashank and Zhelnov, Pavel and Usawasutsakorn, Siranut and Mofayezi, Mohammadreza and Bogdanov, Sergei and Piperski, Alexander and Carauleanu, Marc and Zhang, David K. and Ler, Dylan and Leventov, Roman and Soroko, Ignat and Jansen, Thorben and Lauer, Pascal and Duersch, Joshua and Taamazyan, Vage and Morak, Wiktor and Ma, Wenjie and Held, William and Huy, Tran Đuc and Xian, Ruicheng and Zebaze, Armel Randy and Mohamed, Mohanad and Leser, Julian Noah and Yuan, Michelle X and Yacar, Laila and Lengler, Johannes and Shahrtash, Hossein and Oliveira, Edson and Jackson, Joseph W. and Gonzalez, Daniel Espinosa and Zou, Andy and Chidambaram, Muthu and Manik, Timothy and Haffenden, Hector and Stander, Dashiell and Dasouqi, Ali and Shen, Alexander and Duc, Emilien and Golshani, Bita and Stap, David and Uzhou, Mikalai and Zhidkovskaya, Alina Borisovna and Lewark, Lukas and Vincze, Mátyás and Wehr, Dustin and Tang, Colin and Hossain, Zaki and Phillips, Shaun and Muzhen, Jiang and Ekström, Fredrik and Hammon, Angela and Patel, Oam and Remy, Nicolas and Farhidi, Faraz and Medley, George and Mohammadzadeh, Forough and Peñaflor, Madellene and Kassahun, Haile and Friedrich, Alena and Sparrow, Claire and Sakal, Taom and Dhamane, Omkar and Mirabadi, Ali Khajegili and Hallman, Eric and Battaglia, Mike and Maghsoudimehrabani, Mohammad and Hoang, Hieu and Amit, Alon and Hulbert, Dave and Pereira, Roberto and Weber, Simon and Mensah, Stephen and Andre, Nathan and Peristyy, Anton and Harjadi, Chris and Gupta, Himanshu and Malina, Stephen and Albanie, Samuel and Cai, Will and Mehkary, Mustafa and Reidegeld, Frank and Dick, Anna-Katharina and Friday, Cary and Sidhu, Jasdeep and Kim, Wanyoung and Costa, Mariana and Gurdogan, Hubeyb and Weber, Brian and Kumar, Harsh and Jiang, Tong and Agarwal, Arunim and Ceconello, Chiara and Vaz, Warren S. and Zhuang, Chao and Park, Haon and Tawfeek, Andrew R. and Aggarwal, Daattavya and Kirchhof, Michael and Dai, Linjie and Kim, Evan and Ferret, Johan and Wang, Yuzhou and Yan, Minghao and Burdzy, Krzysztof and Zhang, Lixin and Franca, Antonio and Pham, Diana T. and Loh, Kang Yong and Robinson, Joshua and Gul, Shreen and Chhablani, Gunjan and Du, Zhehang and Cosma, Adrian and White, Colin and Riblet, Robin and Saxena, Prajvi and Votava, Jacob and Vinnikov, Vladimir and Delaney, Ethan and Halasyamani, Shiv and Shahid, Syed M. and Mourrat, Jean-Christophe and Vetoshkin, Lavr and Bacho, Renas and Ginis, Vincent and Maksapetyan, Aleksandr and Rosa, Florencia de la and Li, Xiuyu and Malod, Guillaume and Lang, Leon and Laurendeau, Julien and Adesanya, Fatimah and Portier, Julien and Hollom, Lawrence and Souza, Victor and Zhou, Yuchen Anna and Yalın, Yiğit and Obikoya, Gbenga Daniel and Arnaboldi, Luca and Pokorny), Rai (Michael and Bigi, Filippo and Bacho, Kaniuar and Clavier, Pierre and Recchia, Gabriel and Popescu, Mara and Shulga, Nikita and Tanwie, Ngefor Mildred and Lux, Thomas C.H. and Rank, Ben and Ni, Colin and Yakimchyk, Alesia and Liu, Huanxu (Quinn) and Häggström, Olle and Verkama, Emil and Narayan, Himanshu and Gundlach, Hans and Brito-Santana, Leonor and Amaro, Brian and Vajipey, Vivek and Grover, Rynaa and Fan, Yiyang and Silva, Gabriel Poesia Reis e and Xin, Linwei and Kratish, Yosi and Łucki, Jakub and Li, Wen-Ding and Xu, Justin and Scaria, Kevin Joseph and Vargus, Freddie and Habibi, Farzad and Lian, Long (Tony) and Rodolà, Emanuele and Robins, Jules and Cheng, Vincent and Grabb, Declan and Bosio, Ida and Fruhauff, Tony and Akov, Ido and Lo, Eve J. Y. and Qi, Hao and Jiang, Xi and Segev, Ben and Fan, Jingxuan and Martinson, Sarah and Wang, Erik Y. and Hausknecht, Kaylie and Brenner, Michael P. and Mao, Mao and Jiang, Yibo and Zhang, Xinyu and Avagian, David and Scipio, Eshawn Jessica and Siddiqi, Muhammad Rehan and Ragoler, Alon and Tan, Justin and Patil, Deepakkumar and Plecnik, Rebeka and Kirtland, Aaron and Montecillo, Roselynn Grace and Durand, Stephane and Bodur, Omer Faruk and Adoul, Zahra and Zekry, Mohamed and Douville, Guillaume and Karakoc, Ali and Santos, Tania C. B. and Shamseldeen, Samir and Karim, Loukmane and Liakhovitskaia, Anna and Resman, Nate and Farina, Nicholas and Gonzalez, Juan Carlos and Maayan, Gabe and Hoback, Sarah and Pena, Rodrigo De Oliveira and Sherman, Glen and Mariji, Hodjat and Pouriamanesh, Rasoul and Wu, Wentao and Demir, Gözdenur and Mendoza, Sandra and Alarab, Ismail and Cole, Joshua and Ferreira, Danyelle and Johnson, Bryan and Milliron, Hsiaoyun and Safdari, Mohammad and Dai, Liangti and Arthornthurasuk, Siriphan and Pronin, Alexey and Fan, Jing and Ramirez-Trinidad, Angel and Cartwright, Ashley and Pottmaier, Daphiny and Taheri, Omid and Outevsky, David and Stepanic, Stanley and Perry, Samuel and Askew, Luke and Rodríguez, Raúl Adrián Huerta and Dendane, Abdelkader and Ali, Sam and Lorena, Ricardo and Iyer, Krishnamurthy and Salauddin, Sk Md and Islam, Murat and Gonzalez, Juan and Ducey, Josh and Campbell, Russell and Somrak, Maja and Mavroudis, Vasilios and Vergo, Eric and Qin, Juehang and Borbás, Benjámin and Chu, Eric and Lindsey, Jack and Radhakrishnan, Anil and Jallon, Antoine and McInnis, I.M.J. and Hoover, Alex and Möller, Sören and Bian, Song and Lai, John and Patwardhan, Tejal and 
Yue, Summer and Wang, Alexandr and Hendrycks, Dan},
journal={arXiv},
year={2025}
}

@article{fronsdal2024putnamaxiom,
  title={Putnam-AXIOM: A Functional and Static Benchmark for Measuring Higher Level Mathematical Reasoning},
  author={Kai Fronsdal and Aryan Gulati and Brando Miranda and Eric Chen and Emily Xia and Bruno de Moraes Dumont and Sanmi Koyejo},
  journal={NeurIPS 2024 Workshop on MATH-AI},
  year={2024},
  month={October},
  url={https://openreview.net/pdf?id=YXnwlZe0yf},
  note={Published: 09 Oct 2024, Last Modified: 09 Oct 2024},
  license={Apache 2.0}
}

@inproceedings{kamalloo2023limitations,
  title={Limitations of open-domain question answering benchmarks for document-level reasoning},
  author={Kamalloo, Ehsan and Clarke, Charles LA and Rafiei, Davood},
  booktitle={Proceedings of the 46th International ACM SIGIR Conference on Research and Development in Information Retrieval},
  pages={2123--2128},
  year={2023}
}

@misc{jimenez2024swebenchlanguagemodelsresolve,
      title={SWE-bench: Can Language Models Resolve Real-World GitHub Issues?}, 
      author={Carlos E. Jimenez and John Yang and Alexander Wettig and Shunyu Yao and Kexin Pei and Ofir Press and Karthik Narasimhan},
      year={2024},
      eprint={2310.06770},
      archivePrefix={arXiv},
      primaryClass={cs.CL},
      url={https://arxiv.org/abs/2310.06770}, 
}

@misc{patel2024aimeaioptimizationmultiple,
      title={AIME: AI System Optimization via Multiple LLM Evaluators}, 
      author={Bhrij Patel and Souradip Chakraborty and Wesley A. Suttle and Mengdi Wang and Amrit Singh Bedi and Dinesh Manocha},
      year={2024},
      eprint={2410.03131},
      archivePrefix={arXiv},
      primaryClass={cs.AI},
      url={https://arxiv.org/abs/2410.03131}, 
}

@misc{nakano2022webgptbrowserassistedquestionansweringhuman,
      title={WebGPT: Browser-assisted question-answering with human feedback}, 
      author={Reiichiro Nakano and Jacob Hilton and Suchir Balaji and Jeff Wu and Long Ouyang and Christina Kim and Christopher Hesse and Shantanu Jain and Vineet Kosaraju and William Saunders and Xu Jiang and Karl Cobbe and Tyna Eloundou and Gretchen Krueger and Kevin Button and Matthew Knight and Benjamin Chess and John Schulman},
      year={2022},
      eprint={2112.09332},
      archivePrefix={arXiv},
      primaryClass={cs.CL},
      url={https://arxiv.org/abs/2112.09332}, 
}

@misc{ferrag2025llmreasoningautonomousai,
      title={From LLM Reasoning to Autonomous AI Agents: A Comprehensive Review}, 
      author={Mohamed Amine Ferrag and Norbert Tihanyi and Merouane Debbah},
      year={2025},
      eprint={2504.19678},
      archivePrefix={arXiv},
      primaryClass={cs.AI},
      url={https://arxiv.org/abs/2504.19678}, 
}

@misc{li2025searcho1agenticsearchenhancedlarge,
      title={Search-o1: Agentic Search-Enhanced Large Reasoning Models}, 
      author={Xiaoxi Li and Guanting Dong and Jiajie Jin and Yuyao Zhang and Yujia Zhou and Yutao Zhu and Peitian Zhang and Zhicheng Dou},
      year={2025},
      eprint={2501.05366},
      archivePrefix={arXiv},
      primaryClass={cs.AI},
      url={https://arxiv.org/abs/2501.05366}, 
}

@misc{li2025webthinkerempoweringlargereasoning,
      title={WebThinker: Empowering Large Reasoning Models with Deep Research Capability}, 
      author={Xiaoxi Li and Jiajie Jin and Guanting Dong and Hongjin Qian and Yongkang Wu and Ji-Rong Wen and Yutao Zhu and Zhicheng Dou},
      year={2025},
      eprint={2504.21776},
      archivePrefix={arXiv},
      primaryClass={cs.CL},
      url={https://arxiv.org/abs/2504.21776}, 
}

@misc{sun2025zerosearchincentivizesearchcapability,
      title={ZeroSearch: Incentivize the Search Capability of LLMs without Searching}, 
      author={Hao Sun and Zile Qiao and Jiayan Guo and Xuanbo Fan and Yingyan Hou and Yong Jiang and Pengjun Xie and Yan Zhang and Fei Huang and Jingren Zhou},
      year={2025},
      eprint={2505.04588},
      archivePrefix={arXiv},
      primaryClass={cs.CL},
      url={https://arxiv.org/abs/2505.04588}, 
}

@misc{jiang2025s3dontneeddata,
      title={s3: You Don't Need That Much Data to Train a Search Agent via RL}, 
      author={Pengcheng Jiang and Xueqiang Xu and Jiacheng Lin and Jinfeng Xiao and Zifeng Wang and Jimeng Sun and Jiawei Han},
      year={2025},
      eprint={2505.14146},
      archivePrefix={arXiv},
      primaryClass={cs.AI},
      url={https://arxiv.org/abs/2505.14146}, 
}

@misc{li2025websailornavigatingsuperhumanreasoning,
      title={WebSailor: Navigating Super-human Reasoning for Web Agent}, 
      author={Kuan Li and Zhongwang Zhang and Huifeng Yin and Liwen Zhang and Litu Ou and Jialong Wu and Wenbiao Yin and Baixuan Li and Zhengwei Tao and Xinyu Wang and Weizhou Shen and Junkai Zhang and Dingchu Zhang and Xixi Wu and Yong Jiang and Ming Yan and Pengjun Xie and Fei Huang and Jingren Zhou},
      year={2025},
      eprint={2507.02592},
      archivePrefix={arXiv},
      primaryClass={cs.CL},
      url={https://arxiv.org/abs/2507.02592}, 
}

@misc{wei2025browsecomp,
      title={BrowseComp: A Simple Yet Challenging Benchmark for Browsing Agents}, 
      author={Jason Wei and Zhiqing Sun and Spencer Papay and Scott McKinney and Jeffrey Han and Isa Fulford and Hyung Won Chung and Alex Tachard Passos and William Fedus and Amelia Glaese},
      year={2025},
      eprint={2504.12516},
      archivePrefix={arXiv},
      primaryClass={cs.CL},
      url={https://arxiv.org/abs/2504.12516}, 
}

\newpage
\appendix
\section{Prompts}
\label{app:prompt}

\begin{figure*}[h]
    \centering
    \begin{tcolorbox}[
        colframe=blue!60!black,
        colback=blue!5!white,
        title=Prompt for infer answers in \settingTwo and \settingThree,
        rounded corners,
        width=0.9\linewidth,
    ]
    {\small
    You are a literature analysis agent.

    Problem:
    \{\{Instruction\}\}

    Context Source:
    
    • Use the built-in function fetch\_md to list and read local markdown files under \{\{local\_dir\}\}.
    
    • First, call fetch\_md with instance\_id: "\{\{local\_dir\}\}" and no filename to get the file list.
    
    • Then, call fetch\_md with filename to read selected files.

    Guidelines:
    
    • Tool call policy: Within a single assistant turn, call fetch\_md at most once; read only one document per tool call (one filename). If more documents are required, continue in subsequent turns.
    
    • Sufficiency policy: Answers may not be contained in a single document. After reading, assess whether the currently fetched documents are sufficient to derive the correct answer. If uncertain, continue reading more documents in subsequent turns before finalizing.
    
    • Before the final answer, output your step-by-step reasoning process (Reasoning) explaining how the fetched documents support each claim; include citations like [file: xxx.md] inline.
    
    • Do not guess; provide the final result only once you are confident and can justify it with the read materials.
    
    • IMPORTANT: The final line must be exactly: 
    
    \hspace{5mm} Final answer: <your final answer>
}
    \end{tcolorbox}
    \label{fig:prompt-analysis-answer-2-3}
\end{figure*}

\begin{figure*}[h]
    \centering
    \begin{tcolorbox}[
        colframe=blue!60!black,
        colback=blue!5!white,
        title=Prompt for infer answers in \settingOne,
        rounded corners,
        width=0.9\linewidth,
    ]
    {\small
    You are a concept-based reasoning agent.

    Problem:
    \{\{Instruction\}\}

    Concepts:
    \{\{Concepts\}\}

    Output Requirements:
    
    • First, output a section starting with the exact header: Reasoning:

      \hspace{5mm} Provide step-by-step reasoning to support each claim.
      
    • Then provide the final concise conclusion on the last line in the exact format:
    
      \hspace{5mm} Final answer: <your final answer>
}
    \end{tcolorbox}
    \label{fig:prompt-analysis-answer-1}
\end{figure*}

\begin{figure*}[h]
    \centering
    \begin{tcolorbox}[
        colframe=blue!60!black,
        colback=blue!5!white,
        title=Prompt for infer answers in \settingZero,
        rounded corners,
        width=0.9\linewidth,
    ]
    {\small
    You are an agent skilled at solving challenging problems.

    Problem:
    \{\{Instruction\}\}

    Output Requirements:
    
    • First, output a section starting with the exact header: Reasoning:
    
        \hspace{5mm} Provide step-by-step reasoning to support each claim.
      
    • Then provide the final concise conclusion on the last line in the exact format:
    
      \hspace{5mm} Final answer: <your final answer>
}
    \end{tcolorbox}
    \label{fig:prompt-analysis-answer-0}
\end{figure*}

\begin{figure*}[h]
    \centering
    \begin{tcolorbox}[
        colframe=blue!60!black,
        colback=blue!5!white,
        title=Prompt for extract concepts in CoT,
        rounded corners,
        width=0.9\linewidth,
    ]
    {\small
\# Task Steps\\
1.  Concept Identification: Carefully read the Model's Chain-of-Thought (CoT). Identify and extract every single core concept, theorem, or formula that the model explicitly *used* in its reasoning process.\\
2.  Create Used Concept List: List these extracted items as the Used Concepts.\\
3.  Comparison and Calculation: Compare the Used Concepts list against the Ground Truth Concept List (Standard Concepts) to determine:\\
    - True Positives (TP): Concepts that are in Used Concepts AND in Standard Concepts. (Correctly used necessary concepts)\\
    - False Positives (FP): Concepts that are in Used Concepts BUT NOT in Standard Concepts. (Incorrectly used or irrelevant concepts)\\
    - False Negatives (FN): Concepts that are NOT in Used Concepts BUT are in Standard Concepts. (Necessary concepts that were missed or not used)

---\\
\# Input Data\\
1.  Model's Chain-of-Thought (CoT): The step-by-step reasoning generated by the model.\\
2.  Ground Truth Concept List (Standard Concepts): The authoritative list of core concepts, theorems, or formulas that are *necessary* for a correct solution.

\# INPUT

[Model's Chain-of-Thought (CoT)]

\{\{pred\_cot\}\}

[Ground Truth Concept List (Standard Concepts)]

\{\{concepts\}\}

---\\
\# OUTPUT\\
You must output a JSON object with the following structure, followed by your final calculation.
\textbf{```}json
\{
  "Used Concepts": [
    "Concept A",
    "Concept B",
    ...
  ],
  "True Positives (TP)": count,
  "False Positives (FP)": count,
  "False Negatives (FN)": count,
\}
\textbf{```}
}
    \end{tcolorbox}
    \label{fig:extract-concepts}
\end{figure*}

\begin{figure*}[h]
    \centering
    \begin{tcolorbox}[
        colframe=blue!60!black,
        colback=blue!5!white,
        title=Prompt for error cause analysis,
        rounded corners,
        width=0.9\linewidth,
    ]
    {\small
\# Task

Based on the model’s overall reasoning process (pred\_cot) and the concepts it used (pred\_concepts), compare and match them against the ground-truth core concepts (true\_concepts) and reasoning process (true\_cot), determine the cause(s) of the model’s error, and provide a detailed explanation.

---

\# Core Requirements

Error type labeling rules:
When the matching quality is insufficient, the error cause must be selected strictly from the following labels (the characters must match exactly):

- Errors in reasoning process
- Numerical or formal errors
- Missing core concepts
- Errors in understanding or using core concepts
  If multiple error types co-exist, you may combine up to two labels, separated by “/”.

Specific justification:

- You must explicitly point out the concrete issues in pred\_concepts / pred\_cot
- Clearly explain why they cannot form an effective match with true\_concepts
-  Avoid vague statements; the explanation must be verifiable

---

\# INPUT

true\_concepts: \{\{true\_concepts\}\}\\
true\_cot:\{\{true\_cot\}\}\\
pred\_concepts: \{\{pred\_concepts\}\}\\
pred\_cot: \{\{pred\_cot\}\}\\

---

\# OUTPUT

The output must be a valid JSON in the following format. Do not include any extra fields or explanatory text.

\textbf{```}json
\{
  "error reason": "error in reasoning process",
  "specific reason": "xxxxxxxx in pred\_concepts, and the reasoning process of pred\_cot is xxxxxxxx..."
\}
\textbf{```}
}
    \end{tcolorbox}
    \label{fig:prompt-error-case-analysis}
\end{figure*}

\begin{figure*}[h]
    \centering
    \begin{tcolorbox}[
        colframe=blue!60!black,
        colback=blue!5!white,
        title=Prompt for scoring LLM-generated answers,
        rounded corners,
        width=0.9\linewidth,
    ]
    {\small
The evaluation must adhere strictly to the following Rubric, which assesses three dimensions: A) Factual and Logical Correctness, B) Completeness, and C) Reasoning Quality (based on the CoT).

---

[RUBRIC: Correctness Score (0-1)]\\
1 (Excellent)\\
A: Correctness: The answer is fully correct, logically rigorous, and perfectly aligns with the core conclusion of the true answer.\\
B: Completeness: The answer addresses all aspects of the instruction and includes all key information points from the true answer.\\
C: Reasoning: The CoT is clear, well-structured, and correctly and effectively utilizes all necessary concepts.\\
0.75 (Good)\\
A: Correctness: Mostly correct with minor issues that do not affect the core conclusion.\\
B: Completeness: Covers main parts but may omit secondary details.\\
C: Reasoning: Generally correct and understandable but may have skipped steps or imprecision.\\
0.5 (Acceptable)\\
A: Correctness: Captures the core idea but contains noticeable factual errors or logical flaws.\\
B: Completeness: Only covers the core part, omitting multiple important points.\\
C: Reasoning: Understandable but messy, or fails to sufficiently apply key concepts.\\
0.25 (Poor)\\
A: Correctness: Fundamentally incorrect or based on faulty logic.\\
B: Completeness: Severely deviates from instruction requirements with minimal valid information.\\
C: Reasoning: Serious logical fallacies or irrelevant to the problem.\\
0.0 (Unacceptable)\\
A: Correctness: Completely incorrect, irrelevant, or nonsensical.\\
B: Completeness: Does not address the instruction at all.\\
C: Reasoning: Missing or completely erroneous.\\

---

[INPUT DATA]

Instruction (The Problem to Solve):
\{\{instruction\}\}

True Answer (Benchmark Standard):
\{\{true\_answer\}\}

Predicted Answer (Model Output to Evaluate):
\{\{pred\_answer\}\}

Concepts (Knowledge Base):
\{\{concepts\}\}

Chain-of-Thought (Model's Reasoning Process):
\{\{pred\_cot\}\}

---

[EVALUATION TASK]

Analyze the Predicted Answer and Chain-of-Thought against the True Answer and the Instruction. Assign a single score between 0 and 1 based on the provided Rubric.

Please output only the final score in the exact format:

score: <float between 0 and 1>

IMPORTANT: The last line must be exactly "score: <float between 0 and 1>", and there must be no content after this line.
}
    \end{tcolorbox}
    \label{fig:prompt-scoring-answer}
\end{figure*}




\clearpage
\section{Case Analysis}

\subsection{bad case of Claude-Opus-4.1 in \settingZero}
    \paragraph{Instruction}  Determine the time complexity for the proposed algorithm to provably learn a multi-head attention layer $F$ from random labeled examples drawn uniformly from $\{\pm1\}^{k\times d}$. The learning process must operate under the assumption that the ground truth attention and projection matrices $\{\Theta_{i},W_{i}\}_{i=1}^{m}$ satisfy a set of non-degeneracy conditions, and it should output estimates $\{\hat{W}_{i},\hat{\Theta}_{i}\}_{i=1}^{m}$ that result in a multi-head attention layer $\hat{F}$ achieving a mean squared error of at most $(kd)^{-\Omega(m)}$ with a specified success probability of at least $1-\delta$.
    
    \paragraph{Pred CoT } 
    
    Reasoning:
    
    To determine the time complexity for learning a multi-head attention layer, I need to analyze several key components:
    
    1. Problem Structure: We're learning $m$ attention heads, each with matrices $\Theta_i$ and $W_i$. The input space is $\{\pm1\}^{k \times d}$, giving us $(2)^{kd}$ possible inputs.
    
    2. Target Accuracy: The required MSE of $(kd)^{-\Omega(m)}$ is exponentially small in $m$. This extremely high accuracy requirement significantly impacts the complexity.
    
    3. Sample Complexity Analysis:    
    
    - To achieve MSE of $\epsilon = (kd)^{-\Omega(m)}$, standard learning theory suggests we need $O(1/\epsilon^2)$ samples for concentration   
    
    - This translates to $(kd)^{\Omega(m)}$ samples   - The success probability requirement $1-\delta$ adds a $\log(1/\delta)$ factor
    
    4. Computational Cost per Sample:   
    - For each sample, computing multi-head attention involves:      
    - Attention score computation: $O(k^2d)$ per head      
    - Applying attention weights: $O(k^2d)$ per head        
    - Total per sample: $O(mk^2d)$ for all $m$ heads  
    
    5. Learning Algorithm Complexity:    
    - Provable learning algorithms for neural networks often use moment-based or tensor decomposition methods
    - For multi-head structures with $m$ heads, these methods typically require computing $O(m)$-th order statistics    
    - The computational cost scales exponentially with the order of moments/tensors needed  
    
    6. Total Time Complexity:    
    - Number of samples: $(kd)^{\Omega(m)}$    
    - Cost per sample processing: $O(mk^2d)$     
    - Success probability factor: $\log(1/\delta)$   
    - Combined: $(kd)^{O(m)} \cdot \log(1/\delta)$  
    
    The dominant term is $(kd)^{O(m)}$, which arises from both the sample complexity needed for the exponentially small error and the computational requirements of provable learning methods for multi-head structures.  
    
    Final answer: $(kd)^{O(m)} \cdot \log(1/\delta)$
    
    \paragraph{True Concepts}   
    
    Concept 1: A crude, noisy estimate for the sum of projection matrices, $ \sum_{i} W_{i} $, can be obtained by calculating the empirical correlation between the input sequence $ X $ and the output label $ Y $ via the approximation $ \mathbb{E}\left[\frac{1}{k}X^{\top}JY\right] \approx \sum_{i}W_{i} $, where $ J $ is the all-ones matrix.
    
    Concept 2: A convex body that approximates the affine hull of the attention matrices can be constructed by using linear regression to identify input examples $ X $ that induce similar attention patterns across all heads and leveraging these examples to form affine linear constraints that are simultaneously satisfied by all attention matrices $ \Theta_{1},\ldots,\Theta_{m} $.
    
    Concept 3: The estimate of the projection matrix sum can be significantly refined by using the minimum-norm point within the sculpted convex body as a proxy for the average attention matrix. This allows for the identification of examples that induce large-margin (approximately 1-sparse) attention patterns across all heads, which in turn provides highly accurate linear constraints on the sum of projection matrices.
    
    Concept 4: The convex body sculpting process can be re-run using the refined estimate of the projection matrix sum to produce a new, much more accurate convex body $ K^{*} $ that is significantly closer to the true affine hull of the attention matrices.
    
    Concept 5: An epsilon-net containing candidate attention matrices is constructed not over the full $ d \times d $ space, but specifically within the estimated $ m $-dimensional linear span of the true attention matrices $ \{\Theta_i\}_{i=1}^m $. This crucial dimensionality reduction step reduces the search space for the subsequent brute-force step from being exponential in $ d^2 $ to being exponential only in $ m $.
    
    Concept 6: A brute-force search is performed over all $ m $-tuples of candidate matrices from the low-dimensional epsilon-net, with linear regression solved for each. The final selection of the best model is confirmed on a held-out validation set, and standard generalization bounds are used to certify that this selection is correct with probability at least $ 1-\delta $. This validation step introduces a logarithmic dependency on $ 1/\delta $ to the overall runtime.
    
    \paragraph{Answer status} Incorrect
    
    \paragraph{Causes of errors} Method hallucination due to lack of specific domain knowledge
    
    \paragraph{Missing/incorrect content} Omission of core algorithm mechanisms: Key steps such as "Convex Body", "epsilon-net ($\epsilon$-net)", "Linear Span", and "Brute-force search" were not mentioned.
    
    \paragraph{Incorrect reasoning content} Erroneously introduced "standard learning theory" concentration result $O(1/\epsilon^2)$ as the source of complexity; incorrectly used single-sample computational cost $O(mk^2d)$ as the multiplier for total time complexity, neglecting the significant overhead in the algorithm's training/search phase.
    
    \paragraph{Concepts Recall Status}  Unretrieved / completely missing. Failed to retrieve any core algorithm steps, fully missing the algorithmic process based on convex body geometry and subspace enumeration.
    
    \paragraph{Reasoning state} Logic completely deviated.
    
    \paragraph{Correct logic} Calculate time complexity based on constructive algorithms, with the brute-force search step (searching for $m$ matrices in an $m$-dimensional subspace) dominating the complexity, resulting in an exponential term of $O(m^3)$ (from covering number computation or combinatorial search).
    
    \paragraph{Model logic} Used a "statistical learning theory" template, erroneously simplifying it to "sample complexity × single forward pass computation cost", assuming resolution via moment estimation/tensor decomposition, and treating the sample size $(kd)^{\Omega(m)}$ as the dominant term while neglecting the algorithm's search space complexity.

\subsection{Bad Case of DeepSeek-V3.1 in \settingOne}
    \paragraph{Instruction}
    Using the lifted formulation for certifiable Schubert calculus in type A, determine the total number of bilinear equations in the system $S^A(u, v, w_{\circ}w)$ required to test for non-vanishing Schubert coefficients for the specific case in $S_4$ with permutations $u=2143$, $v=3124$, and $w=4132$. In this context, Schubert coefficients are the structure constants in the ring of Schubert polynomials, and the lifted formulation is a method to create a system of polynomial equations whose solutions correspond to the geometric problem of intersecting Schubert varieties.
    
    \paragraph{Pred CoT}  
    Reasoning: 
    
    - The lifted formulation for certifiable Schubert calculus in type A involves creating a system of bilinear equations  $S^A(u, v, w_{\circ}w)$ whose solutions correspond to the intersection of Schubert varieties. 
    
    - The total number of bilinear equations in the system is determined by the sum of the number of descents for each permutation involved:  $u$,  $v$, and  $w_{\circ}w$. 
    
    - For a permutation, the descent set  $Des(w)$ consists of indices  $i$ where  $w(i) u003e w(i+1)$. - Compute descent sets:   
    
    - For  $u = 2143$:  $\text{Des}(u) = \{1, 3\}$, so  $|\text{Des}(u)| = 2$.   
    
    - For  $v = 3124$:  $\text{Des}(v) = \{1\}$, so  $|\text{Des}(v)| = 1$.   
    
    - For  $w_{\circ}w$, first compute  $w_{\circ}$ for  $S_4$: $w_{\circ} = 4321$. Then $w_{\circ}w = w_{\circ}(w(i))$, with $w = 4132$, so $w_{\circ}w = 1423$. $\text{Des}(w_{\circ}w) = \{2\}$, so $|\text{Des}(w_{\circ}w)| = 1$. 
    
    - Total number of bilinear equations = $|\text{Des}(u)| + |\text{Des}(v)| + |\text{Des}(w_{\circ}w)| = 2 + 1 + 1 = 4$.
    
    Final answer: 4
    
    \paragraph{True Concepts}  
    
    Concept 1: Schubert Coefficients are non-negative integers$c_{u,v}^{w}$ that are defined as the structure constants for the multiplication of Schubert polynomials, which form a linear basis in the ring of polynomials$ \mathbb{Z}[x_1, x_2,  \dots]$. 
    
    Concept 2: 'Geometric Interpretation of Schubert Coefficients states that the coefficient $c_{u,v}^{w}$ counts the number of points in the intersection of three generically translated Schubert varieties, expressed as $c_{u,v}^{w}= \# \{X_{u}(E_{ \bullet}) \cap X_{v}(F_{ \bullet}) \cap X_{w_{ \circ}w}(G_{ \bullet}) \}$.
    
    Concept 3: A Schubert Variety $X_w$ is the Zariski closure of a Schubert cell $ \Omega_w$. The problem of finding the Schubert coefficient is equivalent to finding the number of solutions to a system of equations that define membership in an intersection of these varieties.
    
    Concept 4: Stiefel Coordinates are a set of variables used to parameterize the flags within a Schubert cell $\Omega_w^{A}$. These variables form the entries of a matrix $ \chi_{w}^{A}$ which is central to constructing the polynomial system.
    
    Concept 5: The Lifted Formulation is a method that characterizes membership in a Schubert cell using a system of bilinear equations by introducing auxiliary variables (e.g., $ \alpha,  \beta$), avoiding the use of higher-degree determinantal equations.
    
    Concept 6: The Maximal Descent, denoted by $d :=  \max(Des(u)  \cup Des(v)  \cup Des(w))$, determines the number of column vectors to consider from the Stiefel coordinate matrix and sets the range of indices for the equations in the system.
    
    Theorem 1: "The Characterization of Schubert Cell Membership states that a flag $ \Phi_{ \bullet}$ is in the Schubert cell $ \Omega_{w}^{A}(E^{ \bullet})$ if and only if for each $k  \in [d]$, there exist unique coefficients such that a specific linear combination of the flag's basis vectors lies within a specific subspace defined by the permutation $w$, i.e., $g_k  \in E^{w_{k-1}}- E^{w_k}$. This theorem provides the direct rule for converting the geometric containment problem into algebraic equations.

    \paragraph{Answer status} Incorrect (predicted value: 4, true value: 7)
    
    \paragraph{Causes of errors} Hallucinatory reasoning due to lack of core concepts. The model lacked knowledge of the specific construction of "Lifted Formulation" and fabricated an incorrect heuristic rule (number of equations = sum of descent set element counts).
    
    \paragraph{Missing/incorrect content} Omission of constructive details: Did not mention using $d$ (Maximal Descent) to determine the column range of the Stiefel matrix, introduction of auxiliary variables (to avoid high-degree determinants), or the uniqueness constraint on linear combination coefficients in Theorem 1.

    \paragraph{Error} Incorrect formula $\|Des(u)\| + \|Des(v)\| + \|Des(w_{\circ}w)\|$ (erroneous simplification of the algebraic geometric structure).
    
    \paragraph{Concepts Recall Status} Partially retrieved. Mentioned "Lifted Formulation" and "Schubert Variety", but severely omitted:
    
    Concept 4 (Definition of Stiefel coordinate matrix $\chi_w^A$);
    
    Concept 6 (Definition of Maximal Descent $d$);
    
    Theorem 1 (Theorem on algebraic conversion of Schubert cell membership).
    
    \paragraph{Reasoning state} Logical break (premise error). The break occurred between the first and second steps, where the model fabricated an incorrect formula "total number of equations determined by the sum of descent set sizes" instead of deriving based on the actual construction of Lifted Formulation (Stiefel coordinate dimensions, auxiliary variable introduction, rank constraints), replacing algebraic geometric derivation with simple addition.

\subsection{Bad Case of Gemini-3-Pro in \settingThree}
    
    \paragraph{Instruction}
    For the bumblebee theory, if there exists an external coupled matter field $\mathcal{L}_M$, and the $B$-field is $(0, b(r), 0, 0)$, under the static spherically symmetric coordinate system $ds^2 = -A(r) dt^2 + S(r) dr^2 + r^2 d\Omega^2$, then what is the black hole solution $A(r)$?
    
    \paragraph{Pred CoT}  
    The problem asks for the black hole solution component $A(r)$ in the context of the bumblebee theory with an external coupled matter field and a radial bumblebee field $B_\mu = (0, b(r), 0, 0)$.
    
    Based on the file **\"An exact Schwarzschild-like solution in a bumblebee gravity model.md\"** (which corresponds to the work by Casana et al.), the solution is a spherically symmetric vacuum solution (where the potential $V$ ensures the bumblebee field stays at its vacuum expectation value).
    
    Although the explicit derivation in Section III was truncated in the retrieved text, the metric form is revealed in Section IV.C (\"Time delay of light\"). Equation (75) describes the null geodesic condition $ds^2 = 0$ for the spacetime (labeled as equation 36 in the full text) as:
    
    $$-\left(1 - \frac{2M}{r}\right)dt^2 + (1+\ell)\left(1 - \frac{2M}{r}\right)^{-1}dr^2 + r^2 d\phi^2 = 0$$
    
    Comparing this with the static spherically symmetric metric $ds^2 = -A(r) dt^2 + S(r) dr^2 + r^2 d\Omega^2$, we can identify the metric coefficients:$$A(r) = 1 - \frac{2M}{r}, \; S(r) = \frac{1+\ell}{1 - \frac{2M}{r}}$$
    
    where $\ell = \xi b^2$ is the Lorentz-violating parameter derived from the non-minimal coupling $\xi$ and the vacuum expectation value $b^2$.
    
    The temporal component $A(r)$ remains formally identical to the Schwarzschild solution, while the radial component $S(r)$ is modified by the parameter $\ell$.
    
    Reasoning:
    
    The relevant document [file: An exact Schwarzschild-like solution in a bumblebee gravity model.md] discusses a spherically symmetric solution in bumblebee gravity with a radial background field. The text explicitly writes the line element for null geodesics in Eq. (75) as having a $dt^2$ coefficient of $-(1 - 2M/r)$ and a $dr^2$ coefficient involving $(1+\ell)$. This implies that the metric potential $A(r)$, which corresponds to $-g_{tt}$, is $1 - 2M/r$. This result is consistent with the known Casana et al. (2018) solution where the radial bumblebee field modifies only the $g_{rr}$ component of the metric (and $g_{\theta\theta}$ effective potentials in some contexts, but here $A(r)$ is the focus).
    
    Final answer: $A(r) = 1 - \frac{2M}{r}$
    
    \paragraph{True Concepts}  
    
    Theorem 1: bumblebee theory: $$S = \int d^4x \sqrt{-g} \left[ \frac{1}{2\kappa} R + \frac{\xi}{2\kappa} B^\mu B^\nu R_{\mu\nu} - \frac{1}{4} B_{\mu\nu} B^{\mu\nu} \right] - \int d^4x \sqrt{-g} \, \mathcal{L}_M$$
    
    Theorem 2: We consider the matter field to be an electromagnetic field non-minimally coupled with the bumblebee vector field. Its Lagrangian density expression reads as $$\mathcal{L}_M = \frac{1}{2\kappa} \left( F^{\mu\nu}F_{\mu\nu} + \gamma B^\mu B_\mu F^{\alpha\beta} F_{\alpha\beta} \right)$$, Then through Maxwell equation, we can derive $F_{tr} = \sqrt{A(r)S(r)} \phi'(r)$. Building upon this, we solve the modified Maxwell’s equations to obtain, $\phi(r) = \frac{q}{r}$.
    
    Concept 1: For convenience, we set $\ell = \xi b^2$ as Lorentz-violating parameter, we consider a spacelike background $b_\mu$ with the form $b_\mu = (0,\, b_r(r),\, 0,\, 0)$. Utilizing the aforementioned condition $b_\mu b^\mu = b^2 = \text{const}$, we can derive $b_r(r) = b \sqrt{S(r)}$.
    
    Concept 2: By simplifying Einstein equations, we obtain: $[S(r) A(r)]' = 0$, which implies that $S(r) = \dfrac{C_1}{A(r)}$, where $C_1$ is a constant. Similar to the Schwarzschild-like black hole solution in bumblebee gravity, we set the value of $C_1$ to be $1 + \ell$. Solving Einstein equations $\{t,t\}$, we can get $A(r)$. $$\frac{A'(r) S'(r)}{4 A(r) S(r)} - \frac{A''(r)}{2 A(r)} + \frac{A'(r)^2}{4 A(r)^2} + \frac{S'(r)}{r S(r)} + \frac{2 \phi'(r)^2 S(r)}{(2 + \ell)} = 0$$

    \paragraph{Answer status} Incorrect
    
    \paragraph{Causes of errors} Incorrect derivation path due to neglecting key constraints. The model ignored the "external coupled matter field $\mathcal{L}_M$" setting in the prompt, erroneously simplifying the problem to a vacuum solution (non-charged solution) under pure Bumblebee gravity.
    
    \paragraph{Missing/incorrect content}Omission of electromagnetic field contribution: Did not mention the Lagrangian of the non-minimally coupled electromagnetic field or its Maxwell equation solution $\phi(r) = q/r$.
    
    Omission of source terms in Einstein equations: Failed to include the energy-momentum tensor contribution from $\mathcal{L}_M$ when solving for $A(r)$.

    \paragraph{Missing term in the answer formula} The final $A(r)$ lacks the charge-related correction term $\frac{2(1+\ell)q^2}{(2+\ell)r^2}$.
    
    \paragraph{Concepts Recall Status}   Partially retrieved. Mentioned the definition of Lorentz-violating parameter $\ell$, but severely omitted:
    
    The complete action form of Bumblebee theory;
    
    The matter field Lagrangian $\mathcal{L}_M$ and its Maxwell equation solution $\phi(r) = q/r$;
    
    The solution process of Einstein equations including the matter field (derivation of the correction to $A(r)$ from $\phi'(r)$).
    
    \paragraph{Reasoning state} Logical break. The break occurred in the initial phase of understanding the problem premises and selecting formulas. The problem explicitly states the existence of an "external coupled matter field $\mathcal{L}_M$", but the model erroneously claimed the solution is a spherically symmetric vacuum solution, neglected the matter field, and used a charge-free Schwarzschild-like solution, resulting in the metric coefficient missing the charge $q$ contribution.

\end{document}